# A COMPARATIVE STUDY ON MACHINE LEARNING APPROACHES FOR ROCK MASS CLASSIFICATION USING DRILLING DATA

Tom F. Hansen[*,1,2], Georg H. Erharter[1], Zhongqiang Liu[1], Jim Torresen[2]

[1]Norwegian Geotechnical Institute, Sandakerveien 140, 0484 Oslo, Norway

[2]University of Oslo, Department of Informatics, Gaustadalleén 23 B, 0373 Oslo, Norway

[*]Corresponding author (email: tom.frode.hansen@ngi.no)

## HIGHLIGHTS

- Rock mass quality from MWD data in tunnelling is classified using machine learning
- A big dataset spanning 15 tunnels, with ~500 000 drillholes, boosts model reliability
- The research compares and analyses three model approaches comprehensively
- The models predict Q-class and Q-value as examples of rock mass quality metrics
- A tabular-dataset-based ensemble model outperforms image-based CNN models

## ABSTRACT

Current rock engineering design in drill and blast tunnelling primarily relies on engineers' observational assessments. Measure While Drilling (MWD) data, a high-resolution sensor dataset collected during tunnel excavation, is underutilised, mainly serving for geological visualisation. This study aims to automate the translation of MWD data into actionable metrics for rock engineering. It seeks to link data to specific engineering actions, thus providing critical decision support for geological challenges ahead of the tunnel face. Leveraging a large and geologically diverse dataset of ~500,000 drillholes from 15 tunnels, the research introduces models for accurate rock mass quality classification in a real-world tunnelling context. Both conventional machine learning and image-based deep learning are explored to classify MWD data into Q-classes and Q-values—examples of metrics describing the stability of the rock mass—using both tabular- and image data. The results indicate that the K-nearest neighbours algorithm in an ensemble with tree-based models using tabular data, effectively classifies rock mass quality. It achieves a cross-validated balanced accuracy of 0.86 in classifying rock mass into the Q-classes A, B, C, D, E1, E2, and 0.95 for a binary classification with E versus the rest. Classification using a CNN with MWD-images for each blasting round resulted in a balanced accuracy of 0.82 for binary classification. Regressing the Q-value from tabular MWD-data achieved cross-validated R2 and MSE scores of 0.80 and 0.18 for a similar ensemble model as in classification. High performance in regression and classification boosts confidence in automated rock mass assessment. Applying advanced modelling on a unique dataset demonstrates MWD data's value in improving rock mass classification accuracy and advancing data-driven rock engineering design, reducing manual intervention.

## 1 INTRODUCTION

Measure While Drilling (MWD) data is a big, high-resolution sensor dataset collected automatically in numerous drill and blast tunnelling projects worldwide [1]. Sensors on the drilling machines monitor variations in the rock mass. The returned dataset gives a digital signature of the rockmass for all drilled holes ahead of the face. Besides face-seismics [2], the MWD dataset is the only available dataset ahead of the face with such spatial richness. Traditionally, MWD data has been employed to visualise geological changes, assisting face engineers in making informed decisions on geological challenges ahead of the tunnel face based on their analysis of the complex visual data. The challenge investigated in this study is to automatically translate the sensor data to metrics usable for rock engineering design through a data-driven approach. Specifically, it explores the relationship between the MWD-data and the Q-value using machine learning. The Q-value is an example of a summary value for the inherent stability of the rockmass [3], collected regularly in tunnelling and exists in large datasets. The Q-value is a product of six input values, which are subject to noise and subjective interpretation in the same way as other rockmass classification systems. However, few metrics can easily be collected that describe the whole exposed rockmass from a blasting round with numbers other than existing classification systems.

This research aims to establish a reliable relationship between the Q-class and MWD data, demonstrating the potential to derive rock engineering metrics from MWD data. Our objective is to assess the accuracy of predicting rock mass stability, specifically Q-class and Q-value, using MWD data and machine learning (ML) models. Achieving this can enable the prediction of the Q-class ahead of the tunnel face, facilitating improved decisions on advance support and blasting design based on MWD data. The predicted Q-class can also act as a secondary evaluation of the face-engineer mappings of the Q-class after blasting. A well-performing prediction model trained on data from several tunnels can act as a calibrator of the typical subjective variations from different face engineers mapping similar rock masses.

Recent studies have investigated the relationship between MWD-data and rockmass metrics describing the overall stability or soundness of the rock mass in tunnelling, majorly characterised by a rock mass classification approach [4], [5], [6], [7], [8], [9], [10], [11], [12], identifying significant relationships. However, these studies often face limitations in applicability, generalizability, and representativeness for large-scale tunnelling in a data-driven context, including issues like small datasets, subjective human intervention, and methodological weaknesses. Our study builds upon Hansen et al. [13], expanding the dataset and methodology and significantly enhancing the performance.

Machine learning (ML) methods have advanced research across various fields, including medical diagnosis, exoplanet discovery, and protein structure prediction, establishing ML-based scientific discoveries as a mature methodology [14],[15], [16], [17], [18], [19], [20]. This study uses ML models trained on data from 15 geologically diverse tunnels, encompassing data from approximately 500,000 drillholes of 4-6 meters in length, to delineate the relationship between Measurement-While-Drilling (MWD) data and rock mass characteristics. This extensive dataset poses a significant challenge in terms of fitting a model with good performance and substantially enhances the reliability and generalizability of our machine learning models. Previous research indicated that simple statistical models, such as linear regression, fail to describe this relationship [4] accurately [4], [21], in contrast to nonlinear neural networks, which offer enhanced predictive capabilities [9], [22]. Additionally, the prevalent method of visually correlating MWD images with rock mass stability is subjective and prone to ambiguity [23]. We explore three ML approaches to establish this relationship.

- Statistical summary values for each blasting round are used as tabular sample inputs to tabular-based classification ML algorithms (e.g. KNN, Light Gradient Boosting Method, Multi-Layer Perceptron, or linear regression) to train a classifier to correctly classify a full split in Q-classes (A, B, C, D, E1, E2) and various label combinations of Q-classes.
- The same data were employed to train ML-regression algorithms for continuous Q-value prediction.
- MWD values from contour holes are projected on the contour, thus constructing MWD images with localised sensor information. See details in Hansen et. al [13]. These images are inputs to three different deep learning architectures, all with a Convolutional Neural Network (CNN) at their core.

Deep learning models incorporate data fusion principles to integrate different data sources during training. The performance of conventional models significantly depends on the combination of preprocessing, balancing, and training steps. To ensure a reliable ML modelling process, considerable effort is devoted to preventing data leakage [24] and employing robust evaluation techniques, including multiple metrics and cross-validation [25].

## 2 DATASET

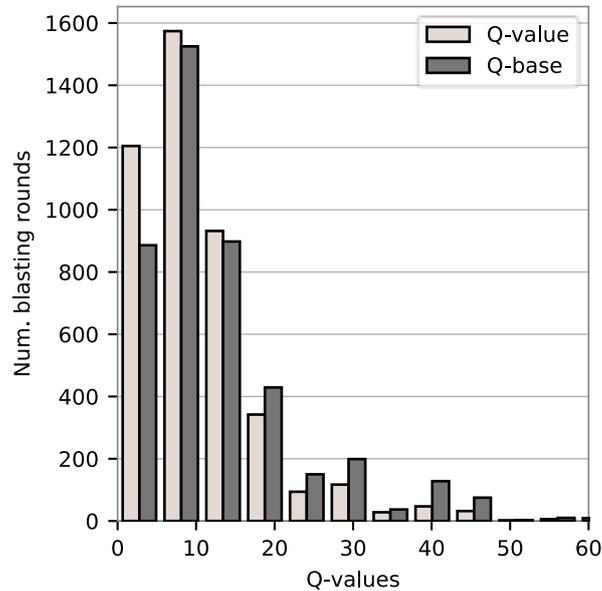

Figure 1. Comparing the distribution of Q-values versus Q-base for single blasting rounds. Notice the shift of values for Q-base to higher values. Twenty samples above 60 are removed from the plot for visual purposes.

The construction and pre-processing of the dataset are described in detail in a separate paper [23]. It contains corresponding MWD-data (features) and rockmass classification in Q-class and Q-value (labels) for 4402 blasting rounds from 15 geologically diversified hard rock tunnels from the four infrastructure projects: UDK, UNB, RV4, E39, in Norway. The main rock types from the tunnels are Precambrian Gneisses, Permian Basalt and Granite, Permian Rhomb porphyry, Cambro-Silurian shales, limestone, and claystone. The MWD parameters used are described in Table 1, and sample sizes are in Table 2. Each of the original MWD-parameter values was first calibrated and then either normalised or Root Mean Squared (RMS) filtered.

For the tabular dataset, all the pre-processed values for all drillholes within a 1 m section of the tunnel (roughly 5 000 sensor values for a standard road tunnel profile of 80 m$^2$ – with 120 drillholes for full face excavation) were used to compute a mean, median, standard deviation, variance, skewness and kurtosis for each of the MWD-parameters, making in total 48 MWD-feature values (6 statistical metrics times eight parameters), and three other geometrical parameters (overburden, tunnel width and Jn-mult). See Figure 2 for a visualisation of the process. We mapped these 51 value-vectors feature samples to the Q-class label given to the blasting round. Given an average blasting length of 5.5 m, we generated 23,277 samples. This method aims to enrich the dataset with additional feature variations under the same Q-class label, which is essential for underrepresented classes. It offers a finer representation of conditions within each 5 m section, aiding the model in identifying more subtle patterns linked to Q-class labels. Dataset augmentation, a well-established machine learning strategy, enhances model performance and generalisability by providing a broader array of examples for training. Future research should examine the model's performance and the effects of varying section lengths. In the image-based dataset, the data from drillholes are projected on the tunnel contour and combined into a matrix of values for each blasting round, visualised as images. Details of the process are described in Hansen et al. [13].

The Q-value is a continuous variable. The spread of values is visualised in Figure 1 with Q-value and Q-base, respectively. Q-base is a reduced version of Q without the last term (Jw/SRF – water presence / Stress Reduction Factor) [26]. The Q-values can be grouped into stability classes from "Exceptionally good", denoted class A, to "Exceptionally poor", denoted class G [27]. Note the distinction between the two ranges in class E, sometimes denoted E1 (Q-value 0.4-1) and E2 (Q-value 0.1-0.4). For Q-values below 0.4, the

Q-system advises using Reinforced Ribs of Shotcrete (RRS) in standard 12 m wide infrastructure tunnels [27]. Infrastructure developers in Norway define specific measures for Q-values under 1.0 (i.e. from class E) [28], [29], which involves more severe actions, such as installing spiling bolts and shorter blasting rounds. Therefore, a binary prediction model that accurately identifies this threshold would have important practical and economic outcomes. Table 2 summarises the number of samples in each class for different grouping levels.

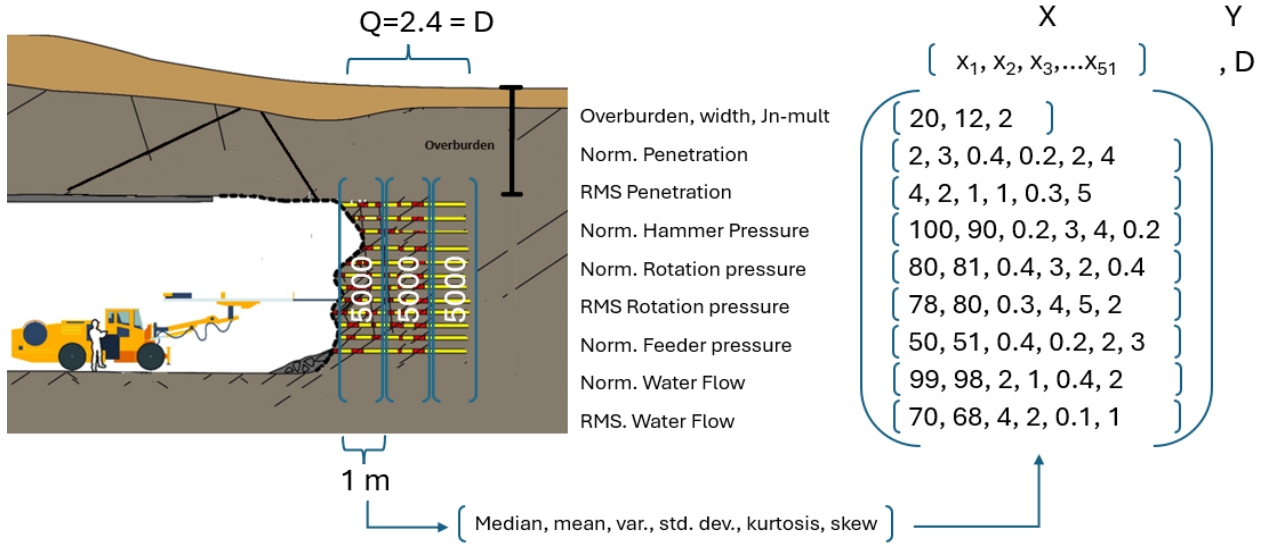

Figure 2. Illustrating the collection of MWD values for 1 m tunnel sections.

Table 1. Features of the dataset used for this study (see [23] for details). Normalised and RMS-filtered MWD-parameters are given by the abbreviated names.

| Type | Original parameter name and unit | Abbreviation for normalised/filtered form in study | Description |
|---|---|---|---|
| MWD-parameter | Penetration rate (m/min) | PenetrNorm | Normalised (norm) penetration, sometimes denoted as Hardness Index. |
| MWD-parameter | Penetration rate (m/min) | PenetrRMS | The root mean square (RMS) filtered penetration value is sometimes denoted as Fracture index. |
| MWD-parameter | Rotation pressure (bar) | RotaPressNorm | Norm. of rotation pressure |
| MWD-parameter | Rotation pressure (bar) | RotaPressRMS | RMS of rotation pressure |
| MWD-parameter | Feeder pressure (bar) | FeedPressNorm | Norm. of feeder pressure |
| MWD-parameter | Hammer pressure (bar) | HammerPressNorm | Norm. of hammer pressure |
| MWD-parameter | Flushing water flow (l/min) | WaterflowNorm | Norm. of waterflow, sometimes denoted as Water Index. |
| MWD-parameter | Flushing water flow (l/min) | WaterFlowRMS | RMS of waterflow |
| Other -/ non-MWD parameter | Overburden (meter) | overburden | Distance from tunnel roof up to surface |
| Other -/ non-MWD parameter | Tunnel width (meter) | tunnel width | Width at the tunnel widest at the base |
| Other -/ non-MWD parameter | Jn-mult | Jn-mult | The parameter is used to calculate the Q-value, which can be determined in advance. Values are set based on the distance to tunnel openings and junctions [27]. |

Table 2. Datasets of grouped labels for Q-class. Numbers for tabular and image datasets.

| Grouped dataset | Samples/rounds in each class tabular dataset | Total num. samples | 1 m - samples in each class tabular dataset | Tot num. samples | Samples in each class image dataset | Total num. samples |
|---|---|---|---|---|---|---|
| A, B, C, D, E1, E2 | A:101, B:1818, C:1712, D:537, E1:167, E2:67, | 4,402 | A:539, B:10057, C:9208, D:2571, E1:642, E2:260, | 23,277 | A:101, B:1640, C:1345, D:404, E1:149, E2:44, | 3,683 |
| A, B, C, D, E | A:101, B:1818, C:1712, D:537, E:234, | 4,402 | A:539, B:10057, C:9208, D:2571, E:902, | 23,277 | A:101, B:1640, C:1345, D:404, E:193, | 3,683 |
| AB, C, D, E | AB:1919, C:1712, D:537, E:234, | 4,402 | AB:10596, C:9208, D:2571, E:902, | 23,277 | AB:1741, C:1345, D:404, E:193, | 3,683 |
| AB, CD, E | AB:1919, CD:2249, E:234, | 4,402 | AB:10596, CD:11779, E:902, | 23,277 | AB:1741, CD:1749, E:193, | 3,683 |
| ABCDE1, E2 | ABCDE1:4168, E2:234, | 4,402 | ABCDE1:22375, E2:902, | 23,277 | ABCDE1:3490, E2:193, | 3,683 |
| AB, CDE | AB:1919, CDE:2483, | 4,402 | AB:10596, CDE:12681, | 23,277 | AB:1741, CDE:1942, | 3,683 |
| AB, DE | AB:1919, DE:771, | 2,690 | AB:10596, DE:3473, | 14,069 | AB:1741, DE:597, | 2,338 |
| A, C, E | A:101, C:1712, E:234, | 2,047 | A:539, C:9208, E:902, | 10,649 | A:101, C:1345, E:193, | 1,639 |

## 3 METHODOLOGY - ARCHITECTURAL CHOICES IN ML MODELS AND TRAINING PIPELINE

The classification task uses two principles for model-fitting: a CNN-based approach with MWD-images as input and a conventional ML approach with tabular MWD-data as input. The rationale for these two approaches was to investigate if the positioning of sensor data in the blasting profile and the higher richness in data for images affected the classification. In this study, the Q-class and Q-value are target labels. If other target labels are chosen, the same architecture can be utilised by adjusting the number of classes in the output. The process leading up to architectural choices is described in this chapter, while results from ML training and evaluation for the chosen architecture are presented in section 4.

### 3.1 Image-based approach

Transfer learning is a standard technique in computer vision that increases performance and generalizability, especially for smaller datasets [30]. The trained model is then finetuned from a pretrained model (here called the backbone), trained on sufficiently similar images. The backbone of this study is a Resnet50 network [31] trained on the ImageNet [32] database with millions of images. All architectures used a cross-entropy loss function with class weightings to handle unbalanced data. A step function was used to learn rate scheduling. Two different learning rates were used for the pretrained backbone network and the new head network, tailored to the number of classes in this task. A lower learning rate was used to finetune the pretrained weights in the backbone compared to the untrained weights in new layers, which are trained from random values and need longer learning steps. After experimentation, it was found optimal to freeze the backbone weights until epoch ten and then open the whole network for training. The adaptive stochastic gradient descent method, Adam [33], was used for optimisation. An early stopping technique was used to avoid overfitting by stopping the training process when the loss value increased over three epochs. The dataset was split into train and test datasets in a 70/30 ratio with the same relative number of samples for each class in both datasets. A computationally demanding training process led to the manual run of hyperparameter optimisation in several runs of the same locked split (set by seed). To ensure reproducible results in running experiments with different configurations and datasets, all configurations were organised and saved using the YACS (Yet another configuration system) system [34].

As a further development of Hansen et al. (2022)[13], the one-image basis was expanded to a three-image input. For each blasting round of 3-7 meters (see architecture in Figure 3), images of hardness, fracturing, and water index were fed to a CNN model with two different architectures described below. When other data sources are present, occasionally multimodal (datatypes of various modalities such as combining MWD-images and tabular data with overburden and tunnel width) data, there is a need to align and fuse the data sources before the final classifier. It is usually beneficial to utilise all relevant feature data. This study uses the terminology of fusion models described in [35]. It was experimented with different versions of fusion principles, ending with the two architectures described below. Each architecture used a fusion principle to concatenate image data and add the tabular input overburden and tunnel width to the image data. CNN-based models were implemented in Python utilising the PyTorch library [36]. In both architectures, the images were normalised and scaled to a [0,1] range before reshaping from 200x600 pixels to 224 x 224 pixels to fit the input requirements to the backbone of the CNN. The image dataset was augmented [37] with random horizontal and vertical flipping in the training process. The two architectures used different forms of data fusion before the final output of classifier probabilities. Other backbones, such as EfficientNetB4 and Densenet161, were tried but did not prove to work better.

Architecture 1 (see Figure 3) accepted a 9-channel input image, i.e., the three 3-channel MWD-images vertically stacked, facilitating a "vertical" relationship between the three MWD-parameters. The standard 3-channel input of the Resnet50 was modified to accept a 9-channel input image. Downscaling to three channels was alternatively investigated by adding a 9-3 channel convolutional layer with a 1x1 kernel before the ResNet50, but the first version worked better. The standard network layer with a 1000-class output of the Resnet50 was removed, and a new smaller head network was attached with a linear layer of 512 neurons before outputting logits for the number of Q-classes used in the study. Architecture 1 can be denoted as a data-level fusion model [35].

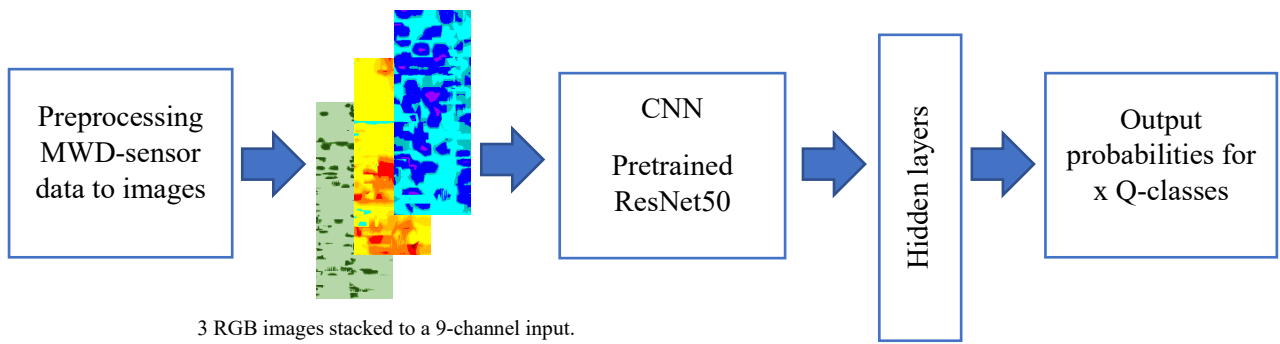

3 RGB images stacked to a 9-channel input.

Figure 3. Architecture 1 – data-level fusion

Architecture 2 (see Figure 4) was designed as three separate CNN pipelines in the core part. The output from each network (num. logits x num. Q-classes) was then concatenated, eventually with some tabular values (normally overburden) and fed into a head network, as for architecture 1. This time, with an input shape of 3 x number of Q-classes + number for overburden. The whole architecture of three distinct CNN pipelines was trained in one operation with standard feed-forward, gradient update and backpropagation. Such an architecture is typically called a feature-level fusion model, given by the concatenations of derived features from each CNN pipeline before entering the final network for classification. Alternatively, it was tried to concatenate feature maps from each CNN pipeline on a higher level (with 2048 output) to preserve more information before the final classification network. Still, the architecture in Figure 4 proved to give better results for this task.

A third fusion type is called decision-level fusion (see Appendix A). An entire CNN pipeline was trained separately and used to output logits (output from the last sigmoid function before being transformed to probabilities using a softmax function) for the six Q-classes for each image (i.e. basically similar to the data-level fusion model). Alternatively, feature maps on a higher level, including 512 outputs, were tried but did not work better. The resulting 18 logits were fed as standard features to a Random Forest algorithm, then outputting the classifications. This architecture often performs best among the fusion principles [35], but not for this model task. Results from this architecture will, therefore, not be further evaluated.

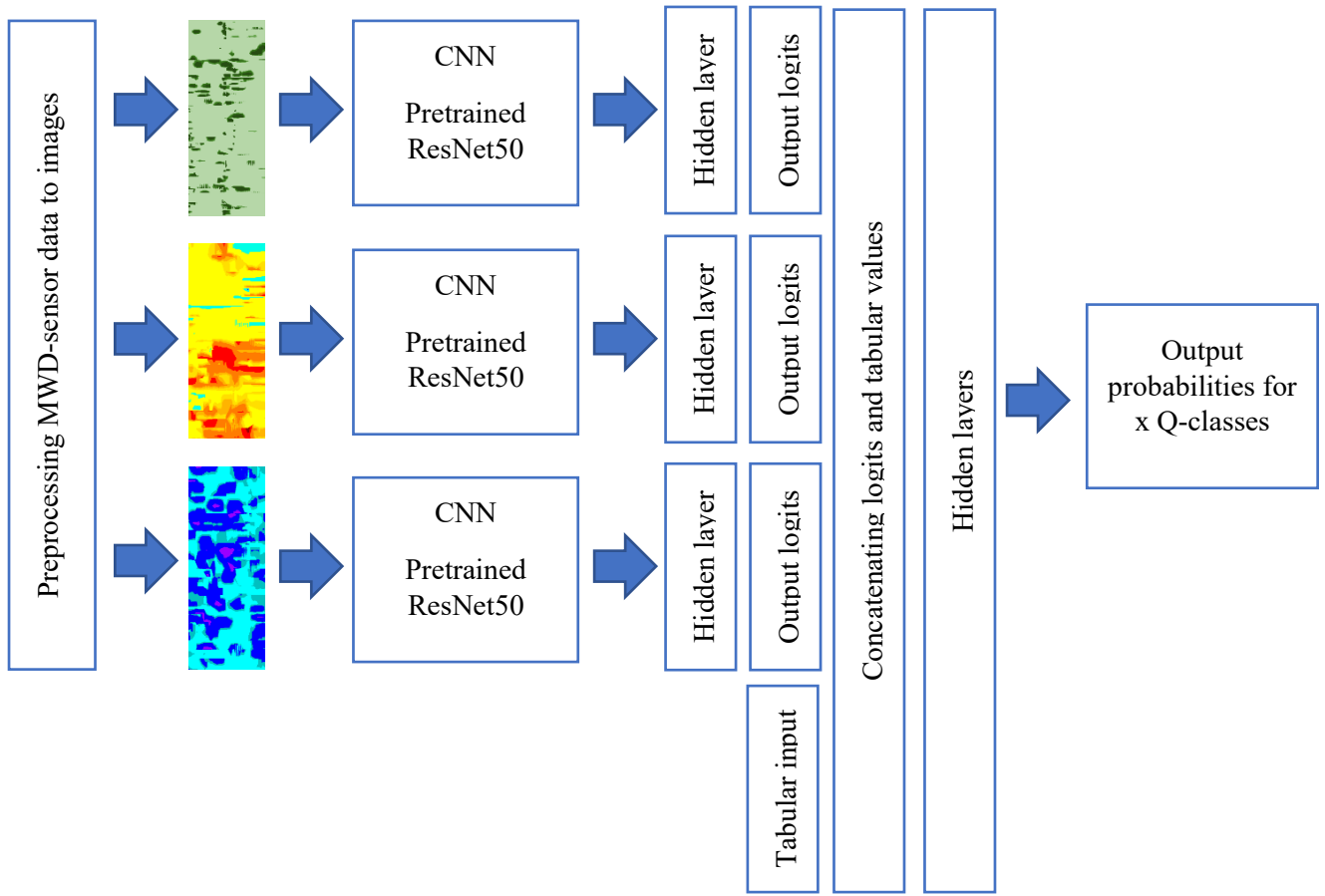

Figure 4. Architecture 2 - feature level fusion.

### 3.2  *Tabular approach*

In the tabular approach, the sample values of features and a Q-class were fed to several classifiers implemented in the Scikit-learn library [38] (Random Forest, Extra trees, KNN, logistic regression, MLP), LGBM [39] and XGBOOST [40]. Even though considerable effort has been put into constructing a balanced and big dataset, it is a natural distribution of more samples in the medium Q-classes. The feature data of the training dataset was balanced using resampling with the SMOTE package [41] before being input to the ML algorithms. Features input to all tested models, besides the tree models, performed best when scaled using a MinMax scaler to the range [0,1]. It was tested to include a PCA step before SMOTE, but this proved ineffective. To avoid data leakage, the process was implemented using the pipeline functionality in Scikit-learn, which ensures the scaling and balancing are only based on the training data. All further training and evaluation used the entire pipeline, described in Figure 5. The training and evaluation followed the process outlined in Hansen et al. [42] to avoid data leakage and ensure a reliable process. The dataset was first splitted into train and test sets in a 75/25 ratio, as recommended by Hastie et al. [43]. Hyperparameters were tuned using Bayesian optimisation with the Optuna package [44]. Each run with a new hyperparameter used 5-fold cross-validation [45] with random splitting of the trainset in train and validation to ensure that the splitting did not severely affect performance. The final training used the whole trainset with optimised parameters before being tested on the untouched testset and running a 5-fold cross-validation process on the full dataset.

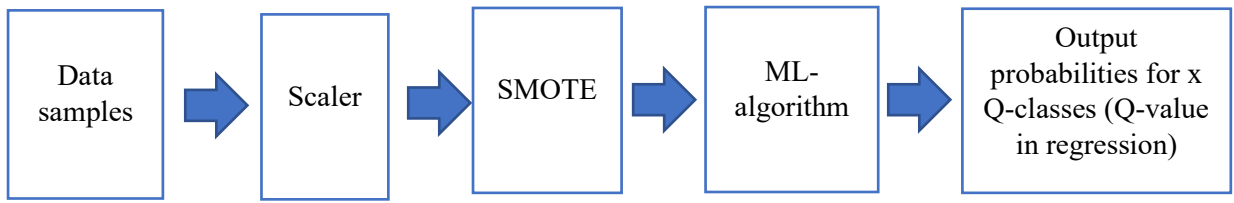

Figure 5. Tabular training process. This process was run to classify Q-classes and regression Q-values. The output for regression was a continuous variable, not probabilities.

Several regression models were fitted to the data, directly investigating the relationship between MWD parameters and the continuous Q-value. Direct regressing avoids the problem of uncertain Q-classes at class boundaries. The model pipeline is identical to the classification, except for the balancing step with SMOTE. Instead of probabilities for Q-classes, a continuous Q-value is the output.

To optimise model performance for regression models, we employed Optuna for hyperparameter tuning. To further enhance our understanding of the model's performance in different ranges of the dataset, we primarily relied on two types of visualisations:

- A scatter plot comparing actual values versus predictions. In an ideal model, the data points would align along the diagonal.
- A Quantile-Quantile (QQ) plot determines if a dataset adheres to a specific theoretical distribution, eventually different distributions in different data ranges, by comparing its quantiles with those of the normal distribution.

We evaluated outlier detection methods for classification and regression, applying univariate (Median Absolute Deviation, MAD) and multivariate (Isolation Forest) techniques and their combinations [46]. To explore the effect of various feature sets, we evaluated models using the following sets:

- A comprehensive set (parameters in Table 1 x six statistical parameters).
- A set omitting independent MWD parameters, feeder- and hammer (percussive) pressure, following Schunnesson [47]. The study described the dependent variables as mainly linked to rock mass characteristics, while the tunnel rig controls independent variables. However, other studies have concluded differently. Navarro et. Al [48] has concluded that feeder pressure is a central parameter related to rock mass characteristics.
- A set automatically reduced using the Featurewiz package [49], which uses the Sulov algorithm for correlated feature elimination and XGBoost for sequential feature selection.
- A set with only MWD parameters, excluding geometric features.
- A set with median values for each MWD parameter.

### 3.3 Metrics for evaluation

For a thorough analysis, reducing the risks of wrong conclusions based on potentially unstable single metrics, we report several metrics [50], [51], [52]. Different metrics report different aspects of performance. In this study, we first address metrics concerning the classification of rock masses into six distinct categories.

**Balanced Accuracy (Average Recall Macro) - Eq. 2:** The primary metric for our model evaluation is Balanced Accuracy, equivalent to the average number of recalls for each class. Recall for a single class is defined as the ratio of true positives (correctly predicted instances of the class) to all actual instances of that class (sum of true positives and false negatives). This metric is particularly relevant in our unbalanced dataset, where standard accuracy might be misleading due to the disproportionate representation of classes. Emphasising balanced accuracy ensures that the bigger number of samples in certain classes does not overly influence our model's performance. This focus is crucial for detecting weaker rock classes, as misclassifying a weak rock as strong could lead to significant safety hazards, such as roof falls or inadequate tunnel face support. Conversely, classifying a strong rock as weak, while not ideal, carries less risk.

**Standard Accuracy -** Eq. 1**:** Our primary focus is on balanced accuracy, but we include standard accuracy as a comparative metric. Standard accuracy measures the proportion of total correct predictions (both true positives and true negatives) out of all predictions made, providing a holistic view of the model's overall performance.

**Average Precision Macro -** Eq. 3**:** We report the Average Macro Precision to complement our focus on recall. Precision for a single class measures the ratio of true positives to the sum of true positives and false positives, indicating the model's ability to correctly identify instances of a class as a proportion of all instances it identified as that class. This metric helps understand the model's performance in terms of false positive rate, which is critical in scenarios where overestimating the strength of a rock mass could be hazardous.

**F1 Score macro -** Eq. 4**:** As a summary metric, we present the F1 macro score, which combines recall and precision. The F1 Score is the harmonic mean of Average Macro Precision and Average Macro Recall, offering a balanced view of both these metrics. It is an essential measure for models that must balance recall and precision. While our primary emphasis is on recall due to the critical nature of correctly identifying weaker rock classes, the lack of consensus in the community on the optimal metric for this task makes it prudent to include the F1 Score. This inclusion allows for a comprehensive evaluation of our models, acknowledging that future insights might reveal the importance of optimising for a metric that equally weighs precision and recall. Future studies can further investigate the optimised precision/recall trade-off for rock mass classification with machine learning.

The **ROC AUC** (Receiver Operating Curve – Area Under the Curve) score is reported to compare different classifiers, a metric that considers both class imbalance and the precision/recall trade-off. The score is calculated using the ROC curve. By including the ROC AUC in our array of metrics, we aim to present a holistic view of our classifiers' performance. This metric complements our primary focus on balanced accuracy and the macro F1 score by offering an aggregate measure of classification effectiveness that remains consistent across various class distributions and threshold settings.

Another important evaluation tool is the **confusion matrix**, giving each class an overview of precision (see Eq. 3) and recall (see Eq. 2). A confusion matrix visualises the unbalances in the classifier and "prediction-leakage" to nearby classes. By normalising the confusion matrix values horizontally and vertically, we return recall values or precision values on the diagonal. As a standard, we have presented the confusion matrix as three matrixes: a recall version, a precision version and a non-normalised version, showing the number of samples predicted correctly or wrongly.

In our regression analysis, we report two key metrics: **$R^2$** (Eq. 5) and **Mean Squared Error** (MSE) (Eq. 6). The R2 metric, also known as the coefficient of determination, provides insight into the proportion of variance in the dependent variable that is predictable from the independent variables, offering a measure of how well unseen samples are likely to be predicted by the model. On the other hand, MSE is used to measure the average of the squares of the errors, i.e., the average squared difference between the estimated values and the actual value, providing a clear and direct indication of the model's prediction accuracy on continuous data.

Balanced accuracy was used as a metric in the objective function in Optuna optimisation for the classification tasks, and the $R^2$-metric was used in regression.

$$\text{Accuracy} = \frac{1}{n} * \sum_{i=1}^{c} (TP\ class\ i\ +\ TN\ class\ i) \quad \text{Eq. 1}$$

TP = True Positives, TN = True Negatives, FN = False Negatives, FP = False Negatives, n = total number of samples

$$\text{Balanced accuracy} = \text{Avg. recall macro} = \frac{1}{C} * \sum_{i=1}^{C} \frac{TP\ \text{class}\ i}{TP\ \text{class}\ i\ +\ FN\ \text{class}\ i} \quad \text{Eq. 2}$$

C is the number of classes

$$\text{Avg. precision macro} = \frac{1}{C} \times \sum_{i=1}^{C} \frac{TP\ \text{class}\ i}{TP\ \text{class}\ i\ +\ FP\ \text{class}\ i} \quad \text{Eq. 3}$$

$$\text{F1 macro} = 2 \times \frac{Avg.\ precision\ macro \times Avg.\ recall\ macro}{Avg.\ precision\ macro + Avg.\ recall\ macro} \quad \text{Eq. 4}$$

$$R^2 = 1 - \frac{\sum_{i=1}^{n}(y_i - \hat{y}_i)^2}{\sum_{i=1}^{n}(y_i - \bar{y})^2} \quad \text{Eq. 5}$$

$y$ is true values, $\hat{y}$ is predicted value, $\bar{y}$ is mean value

$$\text{Mean Squared Error} = MSE = \frac{1}{n} * \sum_{i=1}^{n}(y_i - \hat{y}_i)^2 \quad \text{Eq. 6}$$

## 4 RESULTS AND DISCUSSION

Three distinct machine learning (ML) methods were utilised to correlate Measurement While Drilling (MWD) feature data with established ground condition metrics:

1. Classification of Q-class through traditional ML models using a tabular dataset.
2. Classification of Q-class via a Convolutional Neural Network (CNN) with an image-based dataset.
3. Regression analysis for Q-value and Q-base value from a tabular dataset.

This chapter compares, explores, and discusses the performance of these models.

### 4.1 Classification of Q-class

We have divided the performance investigation into sections focusing on performance at a model level and then detailing it at a class level.

#### 4.1.1 Performance on a model level – tabular approach

Unless explicitly stated, performance metrics for tabular models refer to a KNN model trained on the most challenging full class split (A, B, C, D, E1, E2) and tested on the test set. Most tables display a cross-validated mean score, including variability across five splits, indicated by minimum and maximum values. The selection of the KNN model is justified by results in Table 5, which are elaborated on later in this section.

For the tabular dataset, we first explored the effect of removing outliers with various techniques by training a KNN model on a dataset with outliers removed. In Table 3, we show the impact of different methods, concluding with no outliers removed. The KNN algorithm is sensitive to outliers, as outliers can influence the nearest neighbour search, but the sensitivity is somewhat reduced by MinMax scaling and using the Manhattan distance metric [53], [54]. This setup, combined with the balanced accuracy metric—sensitive to imbalances in the dataset, ensures that the model is particularly attuned to detecting the critical minority class of weak rocks without being misled by their relative rarity or outlier-like status. This finding implies that the so-called outliers, often considered as noise or anomalies in many datasets, are critical data points in classifying drilling data into strong and weak rocks. These 'outliers'—primarily samples of weak rocks—are fewer in number but hold paramount importance for predictive accuracy. Their inclusion without removal enhances the model's ability to recognise the nuanced differences essential for accurate classification. This underscores the necessity of domain-specific considerations in preprocessing and modelling decisions, especially in fields where rare events carry significant information. Our approach demonstrates the value of directly integrating these 'outliers' into the analysis, thereby challenging conventional outlier removal practices and highlighting the importance of a nuanced understanding of data characteristics in predictive modelling.

Table 3. Performance of models trained using different outlier removal techniques. Sample numbers are numbers after outlier removal.

| Outlier removal method | Num. trainset | Num. testset | Balanced acc. | Balanced acc. min | Balanced acc. max | Acc. | ROC AUC |
|---|---|---|---|---|---|---|---|
| no outlier removal | 17,457 | 5,820 | 0.84 | 0.83 | 0.85 | 0.80 | 0.94 |
| multivariate outlier removal with isolation forest (correlation threshold of 0.5) | 17,244 | 5,748 | 0.83 | 0.82 | 0.84 | 0.80 | 0.93 |
| univariate outlier removal with MAD (threshold of 3.5) | 13,047 | 4,350 | 0.80 | 0.79 | 0.83 | 0.78 | 0.92 |
| multivariate and univariate outlier | 12,881 | 4,294 | 0.80 | 0.76 | 0.84 | 0.78 | 0.92 |

We progressed by assessing various feature sets with the KNN algorithm, with results detailed in Table 4. The performance was identical for the "all features" dataset and the automatically reduced feature set (reduced from 51 to 21 features, see Appendix C), highlighting that feature reduction does not compromise model effectiveness. The reduced set prominently includes geometric features and at least two statistical derivatives for all MWD parameters, with median derivatives being the most common. This diversity in derivatives underscores their relevance for each MWD parameter. However, a comprehensive analysis of the significance of each MWD parameter for rock mass prediction is beyond this paper's scope. Our findings also indicate that the "domain features" set, which excludes logically correlated features (e.g., mean and standard deviation), achieves comparable performance, suggesting that removing statistical redundancies like standard deviation versus variance does not enhance model accuracy. Eliminating feeder and hammer pressure decreases the balanced accuracy from 0.84 to 0.82, while excluding geometric features and retaining only MWD parameters lowers it to 0.80. While including geometric features increases the performance of 0.04, the result shows that it is the MWD parameters that contribute almost entirely to the performance, underscoring the potential of correlating rock mass metrics with MWD data. Reducing to just eight median MWD parameters yields a performance of 0.71. In Appendix D, we have visualised confusion matrixes for the four best feature sets to demonstrate the low level of difference among the sets on a class level.

In future applications, selecting the smallest effective feature set, termed "automated features," could simplify the model and improve its explainability. Nonetheless, for comprehensive algorithm comparisons, particularly for detecting weaker rock masses with limited samples, we utilised the "all features" set to ensure a thorough dataset representation in this study.

Table 4. Performance of model trained on different feature sets.

| Feature set | Num. features | Balanced acc. | Balanced acc. min | Balanced acc. Max | Acc. | ROC AUC |
|---|---|---|---|---|---|---|
| Domain features (mean and standard deviation are removed) | 35 | 0.84 | 0.83 | 0.85 | 0.80 | 0.94 |
| All features | 51 | 0.84 | 0.83 | 0.85 | 0.80 | 0.94 |
| Automated features with Featurewiz. A list of features can be found in Appendix C. | 21 | 0.84 | 0.83 | 0.85 | 0.81 | 0.94 |
| MWD-dependent features (feeder pressure and hammer pressure are removed) | 39 | 0.82 | 0.81 | 0.84 | 0.79 | 0.93 |
| MWD features only | 48 | 0.80 | 0.79 | 0.80 | 0.75 | 0.92 |
| MWD median features only | 8 | 0.71 | 0.70 | 0.73 | 0.68 | 0.87 |

In our study, we utilised Optuna for hyperparameter optimisation across different algorithms and scalers, focusing on balanced accuracy as the performance metric representing the mean recall across all classes. Appendix B – figure 1 illustrates the KNN model's performance enhancement through a history plot of various trials, with the red point indicating the outcome using default parameters. The parallel coordinate plot in Appendix B – figure 2 effectively demonstrates the parameter combinations that yield the highest performance. Table 5 showcases the comparative performance of optimised models on a tabular dataset. A dummy model, based on majority voting, achieved a balanced accuracy of 0.17 and an overall accuracy of 0.43. Effective models must outperform this baseline. The KNN model emerged as the top performer in terms of balanced accuracy, likely due to the problem's nature, where similar ground conditions tend to cluster, and KNN discriminates based on the values of nearby points. Tree-based ensemble models also performed well, with the Extra Trees model slightly outperforming others, particularly in average precision. Unlike Random Forest, Extra Trees randomises the split process, reducing overfitting and computational costs, thus enhancing performance in this study [55].

The F1 score, the harmonic mean of precision and recall, benefits from the higher precision of tree-based models compared to KNN despite a slightly lower recall. This suggests that while KNN maintains equal classification across classes, indicated by balanced accuracy, tree-based models like Extra Trees offer a superior precision-recall balance, enhancing the F1 score. The decision to prioritise recall in the classifier pipeline stems from the aim to minimise false negatives, such as misclassifying class E2 as B, deeming it more critical than misclassifying class B as E2. Future research may adopt different optimisation strategies for various rock mass classes, potentially emphasising recall for weaker classes and precision for stronger ones. Further exploration into adjusting precision and recall per class could yield insights into optimal strategies for different rock mass categories.

Table 5. Comparing performance for different algorithms.

| Algorithm | Balanced acc. | Balanced acc. min | Balanced acc. max | Balanced acc. train | Acc. | Avg. precision | F1 | ROC AUC |
|---|---|---|---|---|---|---|---|---|
| Voting Classifier (KNN, Extra Trees, Catboost) – recall optimised | 0.86 | 0.84 | 0.87 | 1 | 0.85 | 0.78 | 0.81 | 0.98 |
| Voting Classifier (KNN, Extra Trees, Catboost) – precision optimised | 0.83 | 0.80 | 0.85 | 1 | 0.86 | 0.83 | 0.83 | 0.98 |
| KNeighbors Classifier | 0.84 | 0.84 | 0.85 | 0.99 | 0.81 | 0.71 | 0.76 | 0.94 |
| Extra Trees Classifier. | 0.79 | 0.77 | 0.81 | 1 | 0.82 | 0.80 | 0.79 | 0.97 |
| CatBoost Classifier | 0.78 | 0.76 | 0.79 | 0.96 | 0.79 | 0.75 | 0.76 | 0.97 |
| XGB Classifier | 0.77 | 0.75 | 0.79 | 0.98 | 0.78 | 0.74 | 0.76 | 0.96 |
| Random Forest Classifier | 0.77 | 0.75 | 0.78 | 0.98 | 0.80 | 0.77 | 0.77 | 0.97 |
| LGBM Classifier | 0.76 | 0.73 | 0.78 | 0.94 | 0.75 | 0.71 | 0.73 | 0.96 |
| MLP Classifier | 0.65 | 0.63 | 0.67 | 0.91 | 0.66 | 0.59 | 0.62 | 0.92 |
| Logistic Regression | 0.51 | 0.49 | 0.52 | 0.53 | 0.40 | 0.32 | 0.32 | 0.83 |
| Decision Tree Classifier | 0.41 | 0.37 | 0.43 | 0.96 | 0.45 | 0.32 | 0.34 | 0.64 |
| Dummy Classifier | 0.17 | 0.17 | 0.17 | 0.17 | 0.43 | 0.07 | 0.10 | 0.50 |

Performance declines when shifting from ensemble methods to single estimators, with neural networks (MLP), Logistic Regression, and Decision Trees ranking lowest. The logistic regression model demonstrates difficulty in fitting the training set, evidenced by a training score of 0.53. Conversely, ensemble models combining various algorithms (voting classifier) excel, offering the best of both worlds by enhancing recall and precision. This approach improves balanced accuracy to 0.86, up from 0.84 with KNN, while maintaining the precision score of Extra Trees. Each algorithm and its scaler are trained separately in a voting classifier before a majority vote determines the final classification. By experimenting with different ensemble mixes, we opted for KNN, Extra Trees, and CatBoost, leveraging their distinct advantages: KNN's non-parametric, lazy learning approach and both bagging and boosting methods.

Additionally, we trained a voting classifier with these algorithms, optimising hyperparameters for precision, thus achieving superior precision and F1 scores. If a single model is preferred for simplicity in interpretation and explanation, KNN is recommended when prioritising recall and Extra Trees are recommended when prioritising precision.

Evaluating the Voting Classifier models' performance by consolidating Q-classes into fewer categories revealed varying impacts on balanced accuracy. See results in Table 6. The benchmark model, with all classes separated, was used as a reference. Initially, reducing the number of classes to four did not enhance performance. However, consolidating into three categories—labelled as "good," "medium," and "bad," akin to a traffic light system—improved balanced accuracy from 0.86 to 0.90. The highest performance was achieved with a binary classification of strong vs. weak, with class E as the minor category, reaching a balanced accuracy of 0.95. Precision showed less variation across different groupings, but a notable improvement from 0.85 to 0.90 was observed when transitioning from a full split to a binary classification. In an additional experiment, we masked intermediate classes B and D to assess the impact of misclassification at class boundaries. Q-class label variability causes border samples to be misclassified by chance, a challenge not entirely solvable by ML models. This experiment demonstrated a performance increase from 0.90 to 0.96 when transitioning from an actual 3-class setup to an artificial one, although this improvement does not represent typical model performance. It effectively highlights the impact of borderline samples on error evaluation in a model.

Table 6. Comparing performance for different label groupings – trained with a Voting Classifier.

| Label grouping | Num. labels | Balanced acc. | Balanced acc. min | Balanced acc. max | Avg. precision | F1 | Acc. | ROC AUC |
|---|---|---|---|---|---|---|---|---|
| A, B, C, D, E1, E2 | 6 | 0.86 | 0.84 | 0.87 | 0.85 | 0.78 | 0.81 | 0.98 |
| A, B, C, D, E | 5 | 0.87 | 0.86 | 0.88 | 0.79 | 0.82 | 0.85 | 0.98 |
| AB, B, C, D, E | 4 | 0.86 | 0.84 | 0.87 | 0.83 | 0.84 | 0.86 | 0.97 |
| AB, CD, E | 3 | 0.90 | 0.88 | 0.91 | 0.85 | 0.87 | 0.89 | 0.98 |
| A, C, E | 3 | 0.96 | 0.96 | 0.98 | 0.94 | 0.95 | 0.98 | 1.00 |
| ABCDE1, E2 | 2 | 0.93 | 0.88 | 0.95 | 0.88 | 0.90 | 1.00 | 1.00 |
| ABCD, E | 2 | 0.95 | 0.94 | 0.96 | 0.90 | 0.92 | 0.99 | 0.99 |

### 4.1.2 Performance on a model level – image-based approach

The decision fusion model, as depicted in Appendix A, emerged as the most effective image-based model architecture, although integrating geometric features did not enhance its performance. Consequently, the performance detailed in Table 7 pertains solely to models trained on MWD-images using the decision fusion approach. The feature fusion model in Figure 4 displayed slightly lower performance, followed closely by the data level fusion model in Figure 3. In contrast, when employing only MWD-parameters for the tabular approach, balanced accuracy dropped from 0.84 to 0.80 across the full range of classes. The image model outcomes, derived from a single train-test split with a 70/30 ratio due to computational constraints, generally fell short of the tabular method's results. Yet, the disparity was smaller compared to the binary tabular model, with image models achieving 0.82 versus 0.95 for the tabular model.

Table 7. Performance metrics on the testset for an image-based CNN-based decision fusion model trained on images made of the MWD-features PenetrNorm, PenetrRMS, and WaterFlowRMS.

| **Label grouping** | **Balanced accuracy** | **Accuracy** |
|---|---|---|
| A, B, C, D, E1, E2 | 0.47 | 0.54 |
| A, B, C, D, E | 0.48 | 0.49 |
| AB, C, D, E | 0.65 | 0.64 |
| AB, CD, E | 0.70 | 0.67 |
| AB, DE | 0.80 | 0.85 |
| ABCD, E | 0.82 | 0.94 |

Using a CNN, Hansen et al. [13] reported balanced accuracies of 0.36 and 0.50 for groupings A, B, C, D, E, and AB, CD, and E, respectively. This study demonstrates improved outcomes with balanced accuracies of 0.48 and 0.70 due to the enhanced capacity of a more sophisticated CNN model that processes MWD-images from three MWD-parameters, as opposed to only PenetrNorm's single-image approach. If utilising MWD-images for predictions is advantageous in tunnel operations, the classifications into good, medium, bad, and the binary model for E-class detection offer valuable insights for advance support decisions. However, the CNN and image-based methods did not outperform the tabular approach, contradicting the hypothesis that localised geological patterns, exclusive to MWD images, significantly impact results. Additionally, the higher informational content in images did not translate to superior classification performance. Despite this, confidence in modelling the relationship between MWD-data and rock mass metrics has grown, supported by the feasibility of using tabular and image-based classification methods.

### 4.1.3 Performance on a Class Level

Model performance metrics aside, understanding how accurately the classifier predicts various classes and the distribution of these predictions is crucial despite balanced accuracy's preference for uniform performance across classes. This insight is depicted through confusion matrices (CM) in Figure 6 - Figure 8 and Appendix E, illustrating results from 5-fold cross-validation on the entire dataset, with test sample evaluations from each split merged into a comprehensive CM. The CMs are presented in three formats:

- The rightmost CM in the figures details the number of predictions versus true labels. For instance, in the case of class A in Figure 7 (with a total of 539 true samples), 498 are correctly predicted as A, while 39 and 2 are misclassified as B and C, respectively. Analysing the vertical column for A reveals that 263-28-1-1 samples from classes B, C, D, and E1 are incorrectly predicted to A. When evaluating the numbers in the CM with a small number of misclassifications, consider that these are 1m long MWD samples. It is unlikely that predictions will be incorrect for the usual 3-6 samples that make up the drillhole length from a blasting round.
- Single CMs are often normalised horizontally (our leftmost CM) to showcase class recall—indicating the model's ability to identify each class—on the diagonal. Cells outside the diagonal are shaded from white to red to signify error values between 0 and 0.5, highlighting the severity of misclassifications relative to the target class. This visualisation is critical for evaluating model reliability, particularly in scenarios like this study, where misclassifying distant classes is more damaging.
- For our CMs, we enhance the analysis by normalising in the vertical direction in the middle figure, which provides precision values (the correct fraction of class predictions) on the diagonal. The choice between optimising for precision or recall involves a trade-off. In rock tunnelling, decisions are often safety-critical and recall typically takes precedence over precision. Correctly identifying the E2 class is more crucial than occasional incorrect classifications of class C as E2. Misclassifying E2 as C poses a significantly greater risk. When feasible, prioritising fewer misclassifications of weaker rock classes is essential.

Figure 6 - Figure 8 displays the CMs for a KNN classifier and two voting classifiers optimised for recall (VCR) and precision (VCP). Initial observations reveal a limited spread of misclassifications and minimal "leakage" to non-neighbouring classes. Many misclassifications among neighbouring classes could be attributed to borderline cases, which are challenging to distinguish due to the subjective nature of the human-assessed Q-value for "continuous" rock mass quality. This suggests the model's actual performance might be better than apparent. Comparing the three CMs, VCR demonstrates superior performance in recall, with negligible misclassifications beyond neighbouring classes. The KNN model exhibits weaknesses in precision, especially for class A and lower classes, whereas precision improves with VCR and is notably better for VCP.

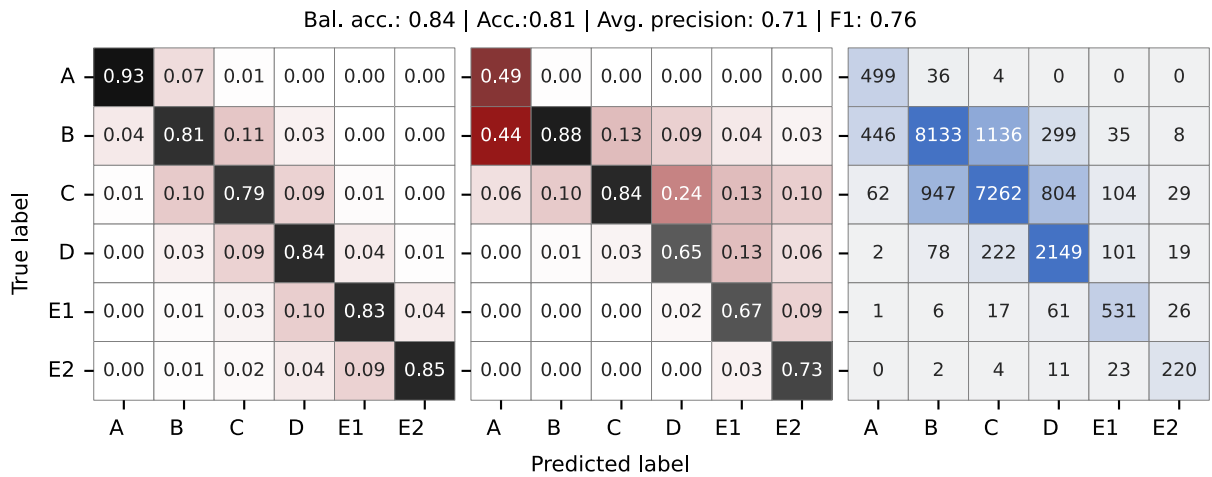

Figure 6. Confusion matrix for a KNN model trained with 5-fold cross-validation.

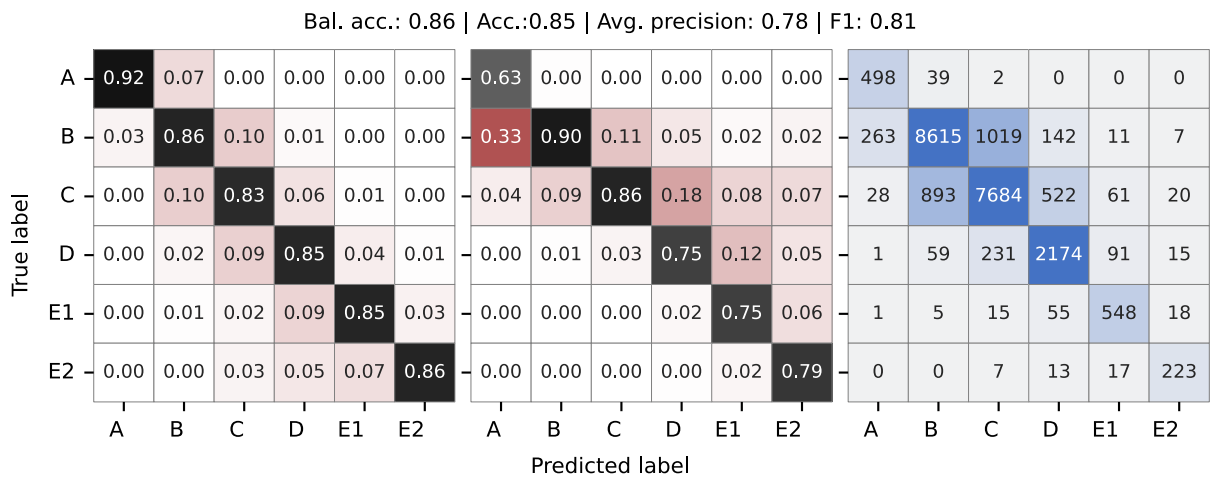

Figure 7. Confusion matrix for a Voting Classifier optimised for recall and trained with 5-fold cross-validation.

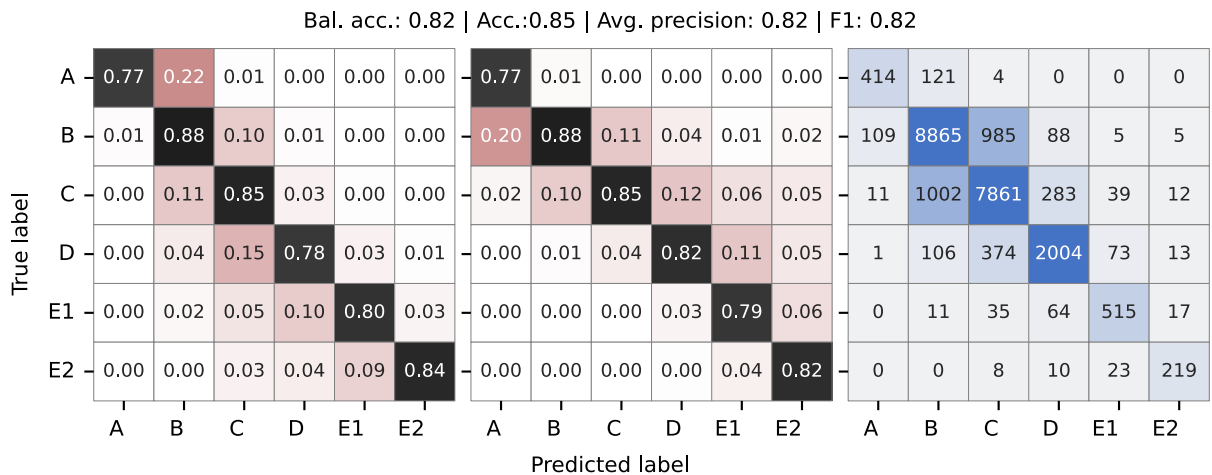

Figure 8. Confusion matrix for a Voting Classifier optimised for precision and trained with 5-fold cross-validation.

The lower precision for classes A, E1, and E2 can be explained primarily by their lower sample sizes. Furthermore, a decrease in sample number tends to result in misclassifications occurring more distantly. Classes with fewer samples face several challenges: inadequate representation for learning variations, increased prediction variability, difficulties in generalisation, and heightened sensitivity to noise [56]. An insight from this pattern is that increasing the sample size will likely reduce these errors, enhancing precision and recall and narrowing the spread of predictions. More samples contribute to improved generalisation and likely reduce errors from incorrect labelling.

Examining the CM for the grouped categories AB, CD, E (Appendix E-1), a significant reduction in misclassification spread is observed, nearing zero, due to the increased "distance" between the now larger classes. This reduction is more pronounced when intermediate classes are masked, as demonstrated in the A, C, E model shown in Appendix E-2. For models aimed at detecting the E-class, which may necessitate heavy support and an extended blasting profile, in a binary classification against a combined ABCD class, the confusion matrix in Appendix E-3 indicates a high balanced accuracy with only an 11% misclassification rate for class E. Thus, a sample classified as ABCD is highly likely to be correct. However, if classified as E, there's a 21% chance it belongs to class ABCD, which is a preferable error from a safety perspective compared to the inverse.

Upon detailed examination of misclassifications, we conducted experiments on a dataset excluding samples from rock mass identified as transition zones, which are areas of noticeable Q-class change. Samples within ten-meter sections (approximately two blasting rounds) following a specific 1 m sample were tagged as transition zones, resulting in 4754 samples classified as such and 18523 as regular. These experiments demonstrated a clear performance improvement: balanced accuracy rose from 0.84 to 0.91 for a VCR model, and average precision increased from 0.78 to 0.85. Conversely, testing exclusively on transition zone samples yielded a balanced accuracy of 0.68 and an average precision of 0.61. The confusion matrices detailing these results are provided in Appendix F. Future research could aim to enhance model performance in transition zones to achieve significant overall improvement.

### *4.2 Regressing Q-values with the tabular dataset.*

Eldert et al. [4] demonstrated a trend by fitting a linear regression line through points of log Fracture index versus log Q across 21 samples. Similarly, Figure 9 displays a log-log plot of Q-value versus PenetrRMS for 4408 samples, marking our initial exploration of simple models to predict the log Q value. The plot reveals a large cluster of points without a distinct trend, and the fitted line poorly explains the data, highlighting the linear regression model's limited predictive capability and the uncertainty involved in fitting a model to a small dataset [53]. Additionally, Figure 1 indicates that the Q-value distribution deviates from normality, a precondition for achieving satisfactory performance with linear models. The low $R^2$ score of 0.31 for multiple linear regression models in this study confirms previous findings regarding the inadequacy of linear models to accurately relate MWD-data to Q-values [5][6]. Consequently, this research has employed more sophisticated ML models in regression analysis to address data complexity more effectively.

We initially optimised hyperparameters for various algorithms combined with scalers using Optuna. Table 8 summarises the cross-validated performance metrics (R2, MSE, and MAE) for regressing log Q-value and log Q-base value. The log transformation of the Q-value label was employed, reflecting the logarithmic nature of the Q-system. Table 8 also includes the MSE score for the training set, demonstrating that, except for multiple linear regression, most models can fit the training dataset perfectly, highlighting the importance of evaluating ML models on the test set.

Consistent with results from Q-class classification, the highest performances among single algorithms were achieved by KNN and Extra Trees, with LGBM replaced by XGBoost among the top three. KNN achieved the best results, with an $R^2$ of 0.80 and an MSE of 0.18. Following the successful approach for classifying Q-class, we fitted an ensemble model using a VotingRegressor, including a pipeline with a scaler for Extra Trees, KNN, and XGBoost, each with optimised hyperparameters. After exploring various combinations in the ensemble, we selected these three, benefiting from their conceptual differences: KNN as a non-parametric lazy learning algorithm, along with a bagging and a boosting method. In the VotingRegressor, outputs from each pipeline were averaged to produce the final value, leveraging the strengths of different models to enhance performance across specific data intervals. This model improved the $R^2$ to 0.81 and reduced the MSE to 0.12. However, given its simplicity, interpretability, and comparable performance, the KNN method is still recommended.

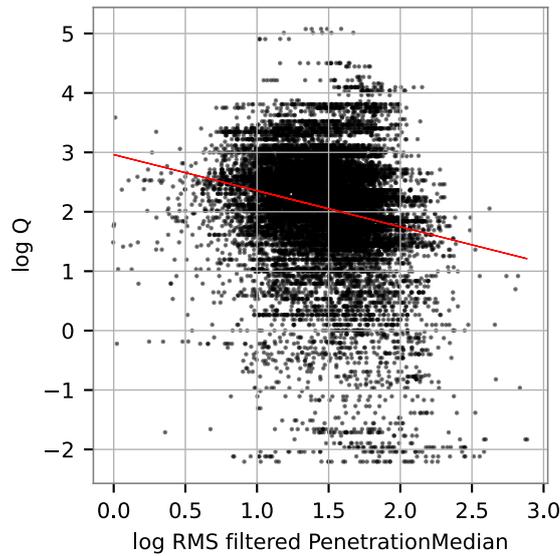

Figure 9. Scatterplot of log PenetrationRMS versus log Q. A linear regression line is fitted to data.

To investigate the potential for reduced noise in labelling, we adapted our model to utilise the Q-base value as the label. This approach involves omitting the Jw/SRF term from the Q-value calculation—a component sometimes contested in literature for its indirect representation of rock mass qualities compared to the initial two terms [57]. This adjustment aims to refine the model's accuracy by focusing on more directly relevant factors of rock mass classification. A model trained using Q-base with a VotingRegressor achieves improved performance, evidenced by an $R^2$ of 0.82 and an MSE of 0.12, making it the highest-performing regression model in our analysis.

All models to date have utilised the complete feature dataset. However, running KNN on subsets featuring only dependent MWD-parameters or exclusively MWD-parameters (excluding geometric features) yielded lower performance, with R2 scores decreasing to 0.76 and 0.73, respectively, as detailed in Table 9. Compared to other studies, recent research employing a similar model setup reported R2 scores ranging from 0.3 to 0.52, attributed to a significantly smaller sample size and outcomes derived from training models on comprehensive data via multiple linear regression [5].

Table 8. Regression metrics for different ML algorithms on regression values for Q-value and Q-base

| Algorithm | Label | R2 | R2 min | R2 max | MSE train | MSE | MAE |
|---|---|---|---|---|---|---|---|
| VotingRegressor(KNN/Extra Trees/XGBoost) | logQ_base | 0.82 | 0.80 | 0.84 | 0 | 0.12 | 0.25 |
| KNeighborsRegressor | logQ_base | 0.81 | 0.80 | 0.83 | 0 | 0.12 | 0.22 |
| VotingRegressor(KNN/Extra Trees/XGBoost) | log Q | 0.80 | 0.78 | 0.82 | 0 | 0.18 | 0.29 |
| KNeighborsRegressor | log Q | 0.80 | 0.78 | 0.82 | 0 | 0.18 | 0.26 |
| XGBRegressor | log Q | 0.77 | 0.74 | 0.78 | 0 | 0.21 | 0.32 |
| ExtraTreesRegressor | log Q | 0.75 | 0.72 | 0.76 | 0 | 0.23 | 0.33 |
| CatBoostRegressor | log Q | 0.73 | 0.70 | 0.74 | 0.14 | 0.25 | 0.36 |
| RandomForestRegressor | log Q | 0.72 | 0.69 | 0.73 | 0.04 | 0.26 | 0.35 |
| MLPRegressor | log Q | 0.57 | 0.49 | 0.61 | 0.29 | 0.39 | 0.45 |
| DecisionTreeRegressor | log Q | 0.37 | 0.30 | 0.46 | 0 | 0.57 | 0.46 |
| LinearRegression | log Q | 0.31 | 0.28 | 0.33 | 0.62 | 0.63 | 0.58 |
| DummyRegressor | log Q | 0 | 0 | 0 | 0.91 | 0.91 | 0.68 |

Table 9. Regression metrics for KNN trained on various feature sets

| Featureset | Num features | Label | R2 | R2 min | R2 max | MSE train | MSE | MAE |
|---|---|---|---|---|---|---|---|---|
| All features | 51 | log Q | 0.80 | 0.78 | 0.82 | 0 | 0.18 | 0.26 |
| All features except independent MWD-parameters (Feeder- and Hammer pressure) | 39 | log Q | 0.76 | 0.72 | 0.79 | 0 | 0.21 | 0.28 |
| Only MWD parameters | 48 | log Q | 0.73 | 0.70 | 0.75 | 0 | 0.25 | 0.31 |
| Only Median MWD features | 8 | log Q | 0.58 | 0.54 | 0.61 | 0 | 0.38 | 0.39 |

Appendix F presents a scatter plot of predicted versus actual values for the KNN model, delineating predictions for the variables log Q and log Q-base. The identity line, representing a hypothetical perfect prediction, is illustrated in red. Data points that deviate beyond a margin of log 2 from this line are highlighted as outliers in red. The Q-Q plots on the left display the distribution of residuals, with the transformation to log Q-base yielding residuals that align more closely with the theoretical line, suggesting normally distributed errors. Particularly for log Q-base, the Q-Q plot confirms the normality of the residuals post-transformation. Conversely, the Q-Q plot for log Q indicates a deviation from normality, especially at the lower end of Q-values, which the model consistently overestimates. This is evidenced by the concentration of red outliers and the divergence from the line in the lower tail of the Q-Q plot, implying a different underlying distribution for these data points. The scatter plots reveal a more even dispersion of red outliers for log Q-base, whereas, for log Q, the outliers are predominantly located at the lower value range. This pattern of skewed predictions is more pronounced in models other than KNN, as exemplified by the Extra Trees model depicted in Appendix H – figures 1 and 2. In these instances, the skewness extends to underestimations at higher values, which may indicate the presence of an imbalanced dataset.

While imbalanced datasets in classification problems are addressed through metrics like balanced accuracy and techniques such as oversampling with Synthetic Minority Over-sampling Technique (SMOTE) [41] and random undersampling [58], the strategies for addressing such imbalances in regression contexts are less developed and not as commonly implemented [59]. Identifying and mitigating these imbalances is crucial for improving model performance and ensuring reliable predictions across the range of the data. To address the observed imbalances and their effects on our regression models, we explored several corrective strategies:

- Dataset "balancing" was attempted by applying a methodology from the resembler-package [60]. This involved segmenting the features and the continuous response variable into discrete bins. Within each bin, we utilised the classification oversampling technique SMOTE to augment underrepresented classes up to a threshold of 1000 samples. The augmented Q-values and features were then employed as the dependent variable in the regression model training. An incremental enhancement in model performance via this technique is evidenced in Appendix H - figure 2.
- We employed sample weighting during the model training phase, which prioritises minimising errors in underrepresented data ranges by assigning them a higher weight.
- The low and high extremities of the class distribution were oversampled to strengthen the representation of these ranges in the training data.
- To rectify the skew in residuals, we fitted a linear regression line to the predictions on the test set and adjusted the model's predictions accordingly. This corrective measure, particularly for the Extra Trees algorithm, is depicted in Appendix H – figure 3. While effective, this approach can be susceptible to variations in the dataset's distribution, potentially altering the skewness pattern and, thus, the efficacy of the adjustment.

Ultimately, it was observed that selecting an appropriate algorithm inherently resistant to data distribution irregularities and class imbalances was the most effective solution. While the techniques above varied in success across different algorithms, the KNN model displayed robust performance from the outset.

Consequently, applying these "boosting" techniques yielded negligible improvements for KNN, emphasising the inherent suitability of this algorithm for our dataset's characteristics.

*4.3    Comparing the performance of different model approaches*

The different modelling approaches delivered varied performance characteristics and applicabilities.

- **Traditional ML Models for tabular data**: Demonstrated robust performance in Q-class classification, with a nuanced understanding of outlier impacts and feature selection critical for predictive accuracy. A Voting Classifier model – an ensemble of KNN, Extra Trees and Cat boost, alongside feature and outlier management strategies, proved effective, especially when domain-specific considerations are prioritised over conventional data preprocessing norms. The performance reached 0.86 balanced accuracies on the full split of classes and 0.95 for binary strong/weak rock mass.
- **CNN for image-based data**: While innovative, the image-based CNN approach showed significantly lower performance than tabular data models, with a balanced accuracy of 0.48 for full class split and 0.82 for binary classification. Despite this, it offered insights into the potential of image-based features for classification tasks, although it did not surpass the effectiveness of traditional ML approaches in this context.
- **Regression analysis for tabular data**: Focused on predicting Q-values, this approach further leveraged the strengths of ensemble methods, integrating KNN, Extra Trees, and XGBoost models into a Voting Regressor, then achieving $R^2$ and MSE scores of 0.80 and 0.18. It highlighted the potential of combining multiple models to enhance prediction accuracy, particularly when dealing with complex, non-linear relationships inherent in geological data.

The comparative performance of these approaches underscores the importance of selecting the appropriate modelling technique based on the specific characteristics of the dataset and the prediction task at hand. Traditional ML models and regression analysis demonstrated superior performance for tabular data, reflecting geological predictions' complexity and nuanced nature. The CNN-based approach, while innovative, suggests that image-based models may require further refinement to match the predictive capabilities of tabular data models for interpreting MWD-data.

## 5    CONCLUSIONS

This study addresses the challenge of converting high-resolution Measurement While Drilling (MWD) data into actionable rock engineering metrics to improve tunnelling safety and efficiency. Previously, this task was hindered by limitations in data quantity, quality, and modelling techniques. We successfully correlated MWD data with rock mass characterisation metrics, specifically Q-value and Q-class, across a diverse dataset of 23,277 samples from 4,402 blasting rounds in 15 tunnels. These tunnels featured predominantly strong, jointed rock masses, with rock types including Gneisses, volcanic rocks, and Cambro-Silurian Shales and Limestones. The following main conclusions have been drawn:

- By employing and comparing three distinct modelling approaches—traditional machine learning algorithms, convolutional neural networks, and regression analysis—we have shown that MWD data can be effectively used to automatically predict the Q-class and Q-value, thus providing a reliable method for evaluating rock mass stability ahead of the tunnel face.
- For classification, we achieved optimal performance using an ensemble model composed of scalers and the machine learning algorithms KNN, Extra Trees, and Catboost, termed a Voting Classifier. We attained a cross-validated balanced accuracy of 0.86, average precision of 0.78, and F1-score of 0.81 across the dataset divided into six classes (A, B, C, D, E1, E2).
- Misclassifications predominantly occurred between neighbouring classes, with samples assigned to weaker rock classes (D and E) showing high confidence in their classification accuracy, minimising incorrect categorisation into stronger rock classes.
- Models trained on data representing more consistent rock mass quality outperformed those in rapidly changing conditions, with scores of 0.91 and 0.68, respectively.

- A modelling approach employing MWD images trained via a CNN architecture demonstrated a significant relationship for datasets classified into strong/weak rock masses. However, the tabular approach yielded higher accuracy, with improvements of 0.15-0.40, but reduced errors in more defined class groupings, such as binary classifications. Despite the detailed location-oriented information provided by MWD-images, they did not outperform the tabular method. Yet, this approach remains viable for binary classification if this data format is preferred.
- A third approach involved regressing the continuous Q-value from MWD-parameters using an ensemble model similar to the tabular case but substituting CatBoost with XGBoost. This resulted in a strong correlation with $R^2$ values of 0.80 and MSE of 0.18. Enhancing the model to predict the less noisy Q-base value improved performance to $R^2$ values of 0.82 and MSE of 0.12.
- Given the preference for simplicity, attributed to enhanced explainability, interpretability, and fewer data collection errors, employing the Q-base model for regression and a single KNN model for tabular classification is effective, reducing the risk of overfitting.
- This analysis was based on a dataset featuring 48 MWD attributes and three geometric features. Geometric features contributed approximately 0.04 to the overall performance. The model still performs well, excluding geometric features and relying solely on MWD parameters.

The model's applicability varies with the specific use case. For projects aiming to distinguish the rock mass ahead of the tunnel face as strong or weak, a binary classification (ABCD versus E) is recommended to determine the need for heavy support measures. With a recall and precision for class E at 0.89 and 0.79 (full model performance is 0.94 and 0.90), the model's accuracy is considered sufficient for practical application, likely surpassing human assessment capabilities. Notably, while samples predicted as strong are accurately identified, 21% of samples predicted as weak are misclassified, which is considered less critical. Using this model, personnel would be concretely alarmed by potential stability risks ahead, and there is less need to interpret the complex MWD pattern from drillholes ahead by human judgment.

## 6 FUTURE RESEARCH AND PRACTICAL APPLICATIONS IN ROCK MASS ASSESSMENT

The remaining error for the tabular conventional ML model may arise from several factors: noisiness of MWD-data, especially in zones of transitioning rock mass quality, borderline cases wrongly labelled in the human labelling, too few samples in some classes, and the attempt to assign discrete Q-classes to the rock mass. While some of these issues might be optimisable, this underscores the difficulty of achieving 100 % accuracy. When assessing the performance, it is noteworthy that these are based on data from 1 m-long tunnel sections. However, predictions are unlikely to be incorrect for all the 3-6 samples typically representing the drill hole length per blasting round. Predicting in 1m sections offers a more detailed rock mass classification than the current practice of assessing on a blasting round basis.

Future research should examine integrating confidence metrics with classifications to aid face engineers in decision-making. The precision-recall trade-off must be further explored to optimise the use of ML models in rock mass metric prediction, mirroring the approach in medical fields like cancer diagnostics. Expanding the dataset to encompass a broader range of geological conditions and tunnelling environments will enhance model generalisability and facilitate improvements in areas of rapidly changing rock mass conditions. Enhancing the explainability of predictions and model interpretability will bolster trust in its predictions.

Establishing a significant relationship between MWD data and Q-classes across three modelling approaches on a big and diversified dataset validates using MWD data as a reliable indicator of rock mass conditions. It demonstrates the feasibility of data-driven rock mass assessment without manual intervention. With continued model refinement and industry acceptance, such a model, based on thousands of blasting rounds, could serve as a benchmark for resolving rock mass assessment disputes among contractual parties. Future research with big and varied datasets may successfully link MWD data to additional outcomes, leveraging the methodologies and frameworks outlined in this study.

# 7 CREDIT AUTHOR STATEMENT

# 8 ACKNOWLEDGEMENT

The authors gratefully acknowledge Thorvald B. Wetlesen, Ivar Oppen, and Christian H. Svendsen from the tunnel software/hardware company Bever Control, which has helped develop ideas, review, and facilitate the data from the clients Bane NOR, Statens Vegvesen, Nye Veier, and the contractor AF-Gruppen.

This research received no specific grant from funding agencies in the public, commercial, or not-for-profit sectors.

# 9 DECLARATION OF GENERATIVE AI AND AI-ASSISTED TECHNOLOGIES IN THE WRITING PROCESS

While preparing this work, the authors used GPT-4 from OpenAI to improve the readability and language of some paragraphs in the text. After using this tool/service, the authors reviewed and edited the content as needed and take full responsibility for the content of the publication.

# APPENDIX

Appendix A. Decision-level fusion architecture

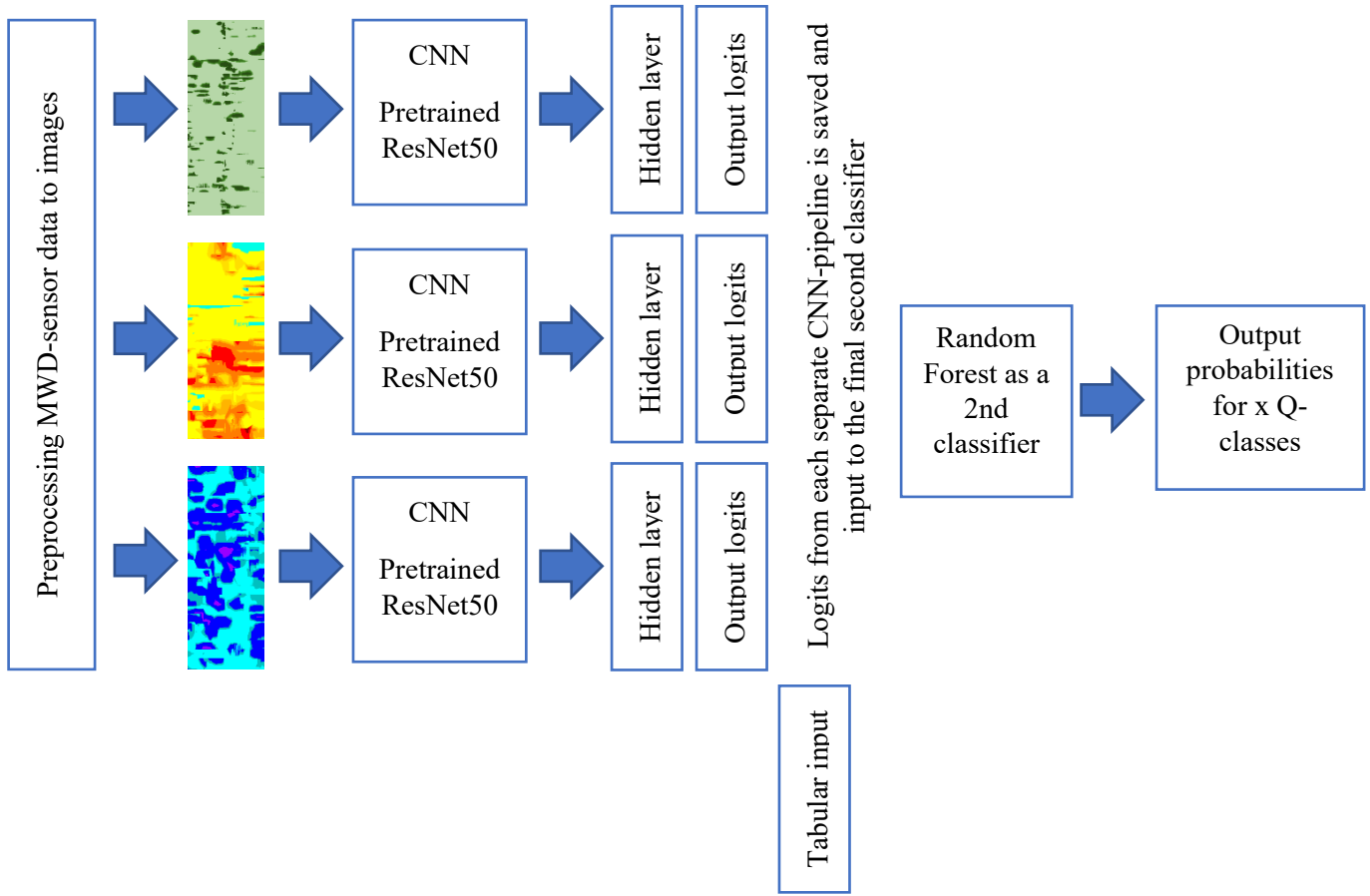

Appendix B. Plots of hyperparameter optimisation history for the KNN algorithm. B1 shows performance for each trial, with a red-coloured dot for the trial using default parameters. B2 shows a parallel coordinate plot of all parameter combinations, highlighting the paths of the combinations of higher performance parameters.

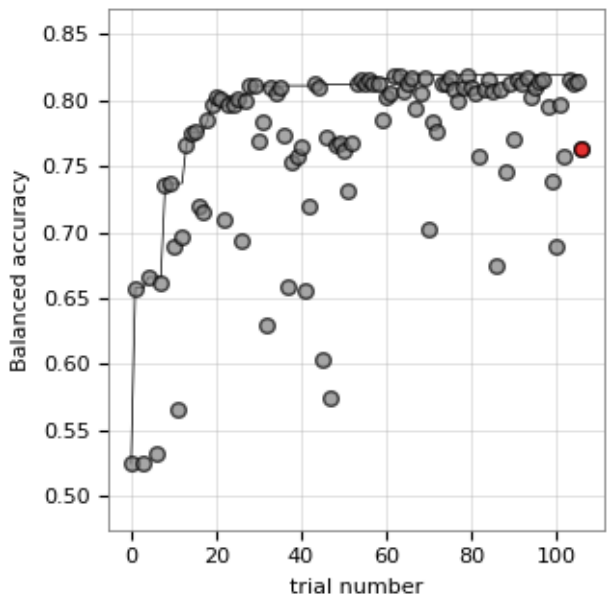

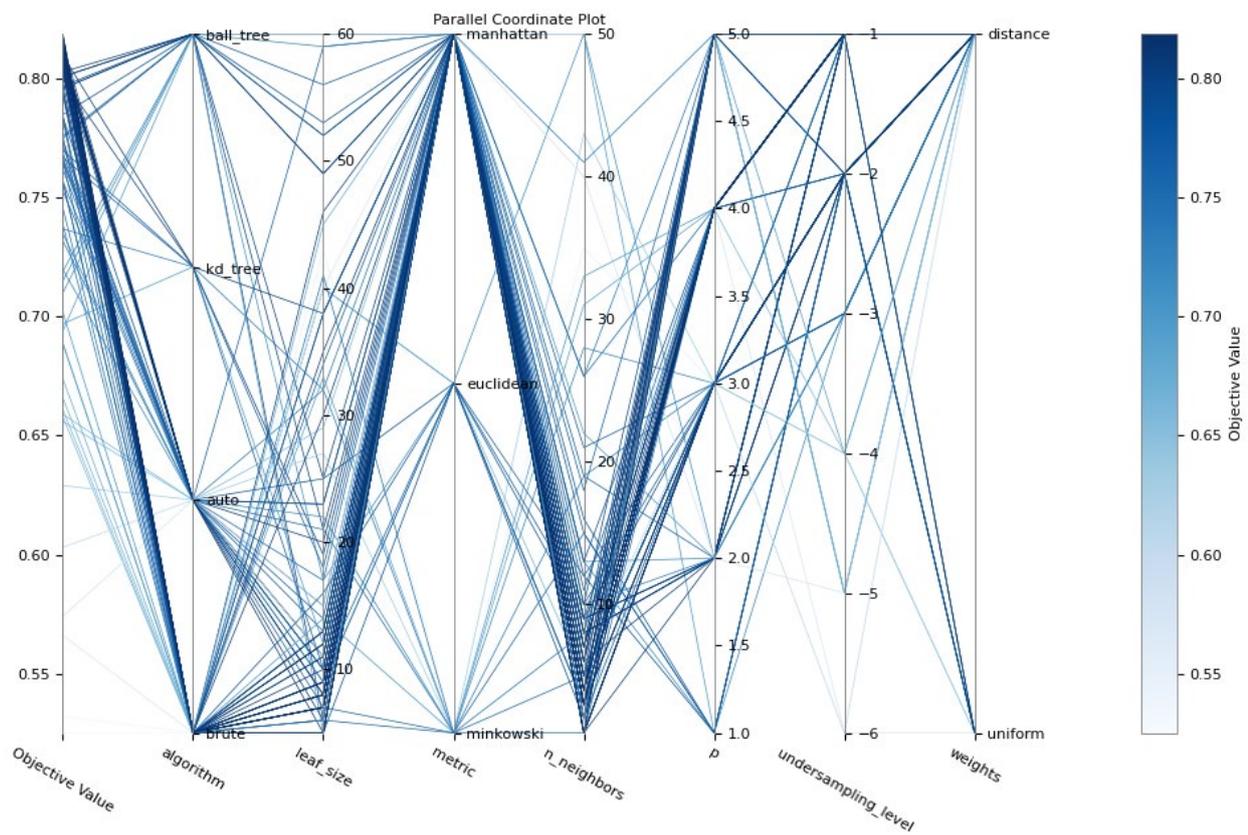

Appendix C. Reduced feature set, automated with Featurewiz.

ContourWidth, TerrainHeight, JnMult, FeedPressNormMedian, FeedPressNormVariance, HammerPressNormMedian, HammerPressNormKurtosis, PenetrNormMedian, PenetrNormStandardDeviation, PenetrRMSMean, PenetrRMSKurtosis, PenetrRMSVariance, RotaPressNormMedian, RotaPressRMSMean, RotaPressNormStandardDeviation, RotaPressRMSKurtosis, RotaPressRMSVariance, WaterFlowNormMedian, WaterFlowNormSkewness, WaterFlowRMSKurtosis, WaterFlowRMSStandardDeviation

Appendix D. Comparing confusion matrixes for a model trained with KNN on the dataset with no label grouping and different feature sets. Top left: complete feature set. Top right: Automatically chosen featureset. Bottom left: MWD-dependent features + geometric features. Bottom right: MWD features only.

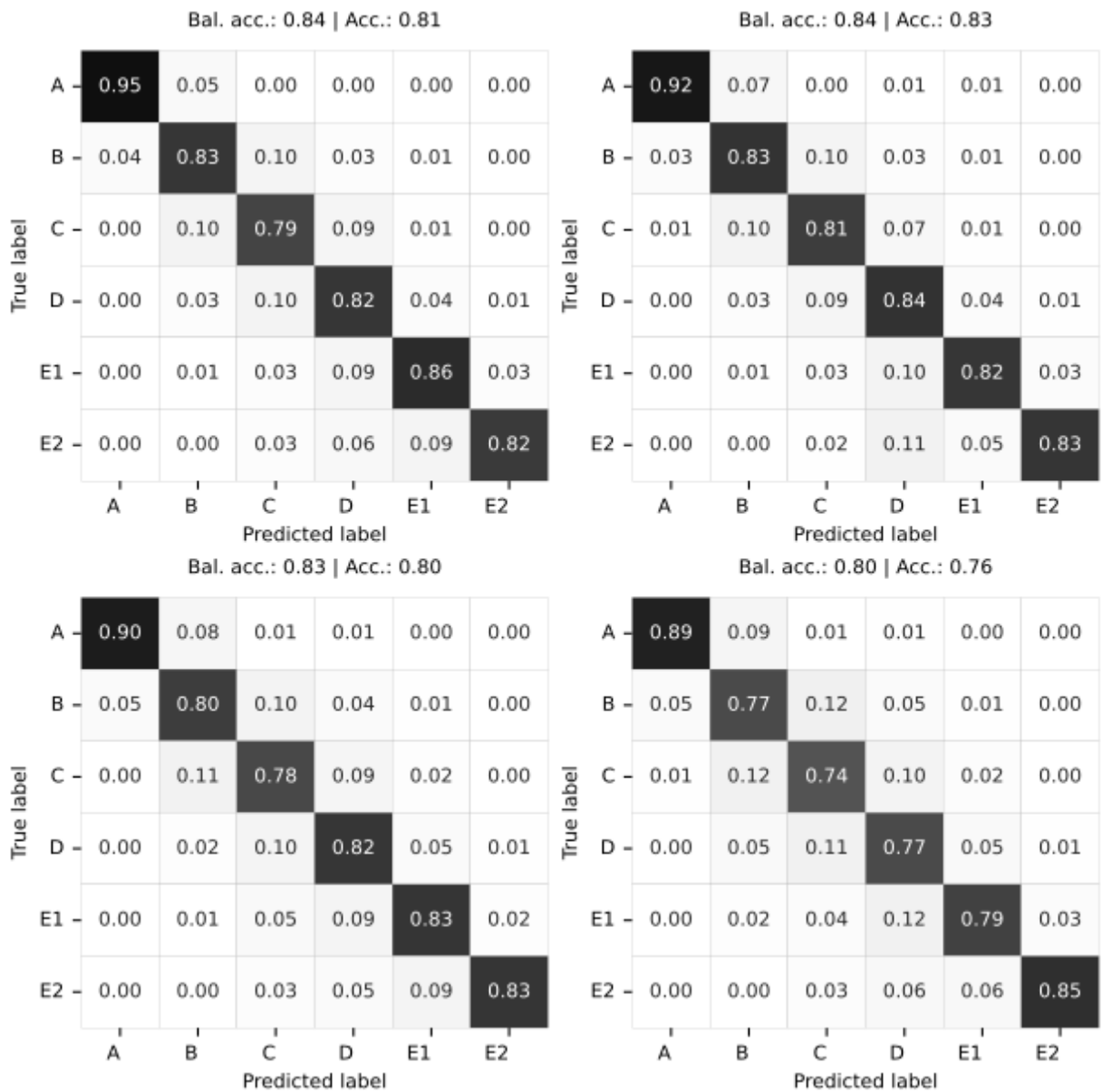

Appendix E. Confusion matrixes for a Voting Classifier optimised for precision and trained with 5-fold cross-validation. E1. Trained on a AB, CD, E grouping. E2. Trained on A, C, E (masked out B and D). E3. Trained on the binary ABCD, E.

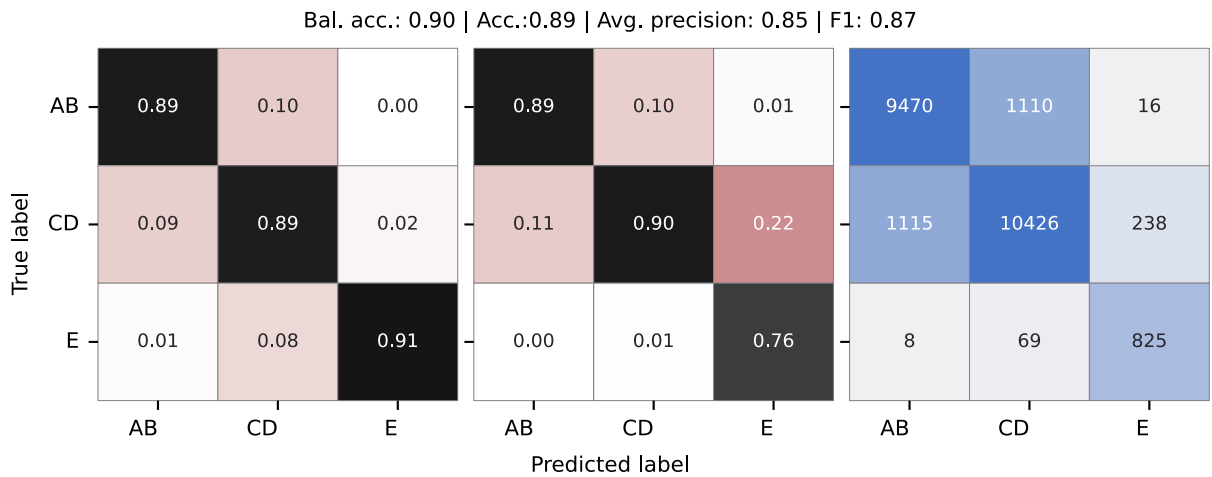

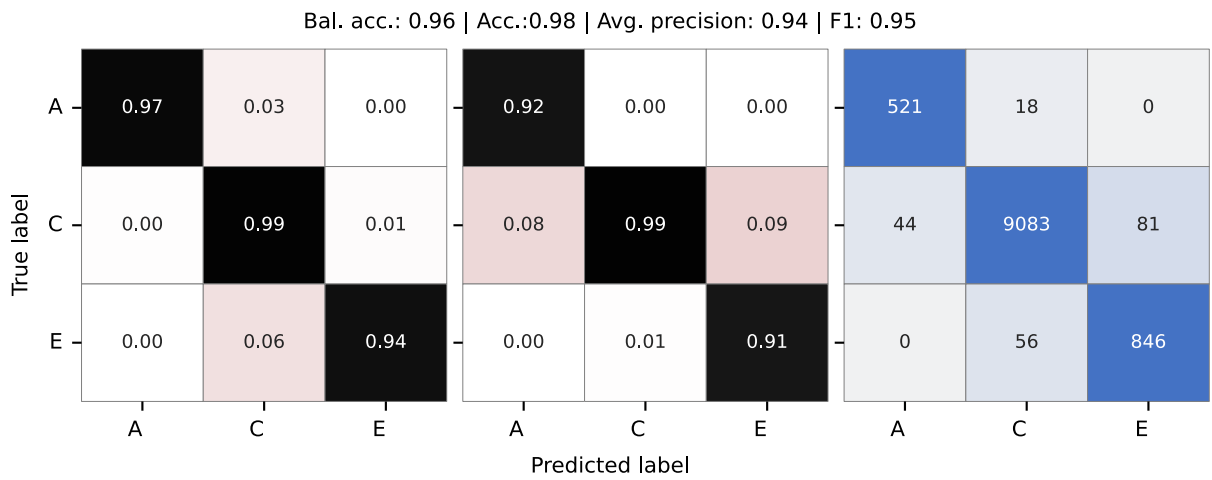

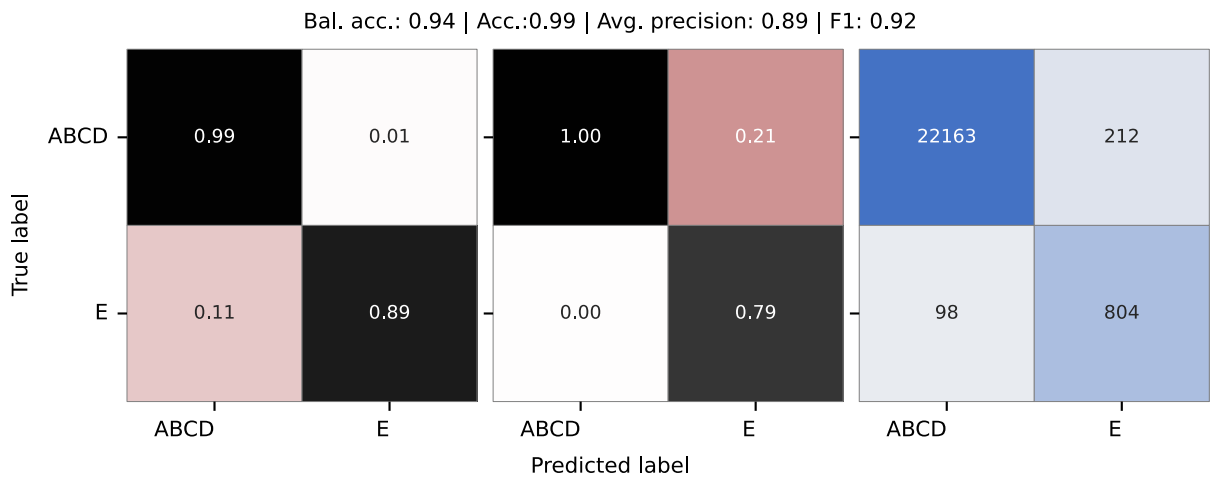

Appendix F. F1. The confusion matrix of a model trained with the voting classifier on the full dataset was only tested on samples in regular zones. F2. Confusion matrix of a model trained with voting classifier on the full dataset and only tested on samples in transition zones.

Bal. acc.: 0.93 | Acc.:0.91 | Avg. precision: 0.84 | F1: 0.88

|  | A | B | C | D | E1 | E2 |  | A | B | C | D | E1 | E2 |  | A | B | C | D | E1 | E2 |
|---|---|---|---|---|---|---|---|---|---|---|---|---|---|---|---|---|---|---|---|---|
| A | 0.95 | 0.05 | 0.00 | 0.00 | 0.00 | 0.00 |  | 0.63 | 0.00 | 0.00 | 0.00 | 0.00 | 0.00 |  | 75 | 4 | 0 | 0 | 0 | 0 |
| B | 0.02 | 0.91 | 0.05 | 0.01 | 0.00 | 0.00 |  | 0.34 | 0.94 | 0.06 | 0.05 | 0.03 | 0.02 |  | 40 | 1936 | 115 | 26 | 3 | 1 |
| C | 0.00 | 0.06 | 0.90 | 0.03 | 0.01 | 0.00 |  | 0.03 | 0.05 | 0.92 | 0.12 | 0.10 | 0.06 |  | 4 | 105 | 1641 | 62 | 11 | 3 |
| D | 0.00 | 0.02 | 0.06 | 0.91 | 0.01 | 0.01 |  | 0.00 | 0.00 | 0.01 | 0.82 | 0.03 | 0.06 |  | 0 | 9 | 26 | 415 | 3 | 3 |
| E1 | 0.00 | 0.01 | 0.00 | 0.02 | 0.97 | 0.00 |  | 0.00 | 0.00 | 0.00 | 0.00 | 0.84 | 0.00 |  | 0 | 1 | 0 | 2 | 97 | 0 |
| E2 | 0.00 | 0.00 | 0.00 | 0.02 | 0.02 | 0.96 |  | 0.00 | 0.00 | 0.00 | 0.00 | 0.01 | 0.86 |  | 0 | 0 | 0 | 1 | 1 | 43 |

Bal. acc.: 0.67 | Acc.:0.62 | Avg. precision: 0.61 | F1: 0.63

|  | A | B | C | D | E1 | E2 |  | A | B | C | D | E1 | E2 |  | A | B | C | D | E1 | E2 |
|---|---|---|---|---|---|---|---|---|---|---|---|---|---|---|---|---|---|---|---|---|
| A | 0.79 | 0.21 | 0.00 | 0.00 | 0.00 | 0.00 |  | 0.65 | 0.03 | 0.00 | 0.00 | 0.00 | 0.00 |  | 44 | 12 | 0 | 0 | 0 | 0 |
| B | 0.06 | 0.60 | 0.30 | 0.03 | 0.01 | 0.01 |  | 0.34 | 0.61 | 0.27 | 0.06 | 0.03 | 0.09 |  | 23 | 237 | 118 | 12 | 2 | 2 |
| C | 0.00 | 0.27 | 0.58 | 0.13 | 0.02 | 0.00 |  | 0.01 | 0.33 | 0.64 | 0.30 | 0.10 | 0.09 |  | 1 | 129 | 274 | 62 | 8 | 2 |
| D | 0.00 | 0.04 | 0.20 | 0.66 | 0.10 | 0.01 |  | 0.00 | 0.02 | 0.09 | 0.59 | 0.25 | 0.05 |  | 0 | 7 | 37 | 123 | 19 | 1 |
| E1 | 0.00 | 0.03 | 0.02 | 0.15 | 0.73 | 0.07 |  | 0.00 | 0.01 | 0.00 | 0.04 | 0.57 | 0.18 |  | 0 | 2 | 1 | 9 | 44 | 4 |
| E2 | 0.00 | 0.00 | 0.05 | 0.10 | 0.20 | 0.65 |  | 0.00 | 0.00 | 0.00 | 0.01 | 0.05 | 0.59 |  | 0 | 0 | 1 | 2 | 4 | 13 |

True label (vertical axis) · Predicted label (horizontal axis)

Appendix G. True labels versus predicted for regression of log Q-values using the model KNN. The left figures show a QQ plot and the right figures show a scatter plot of true values versus predicted values. The two upper figures are for logQ-base, and the lower is for standard log Q. The red line describes values for a perfect model. Predictions outside a band of 2 are defined as outliers and are coloured red.

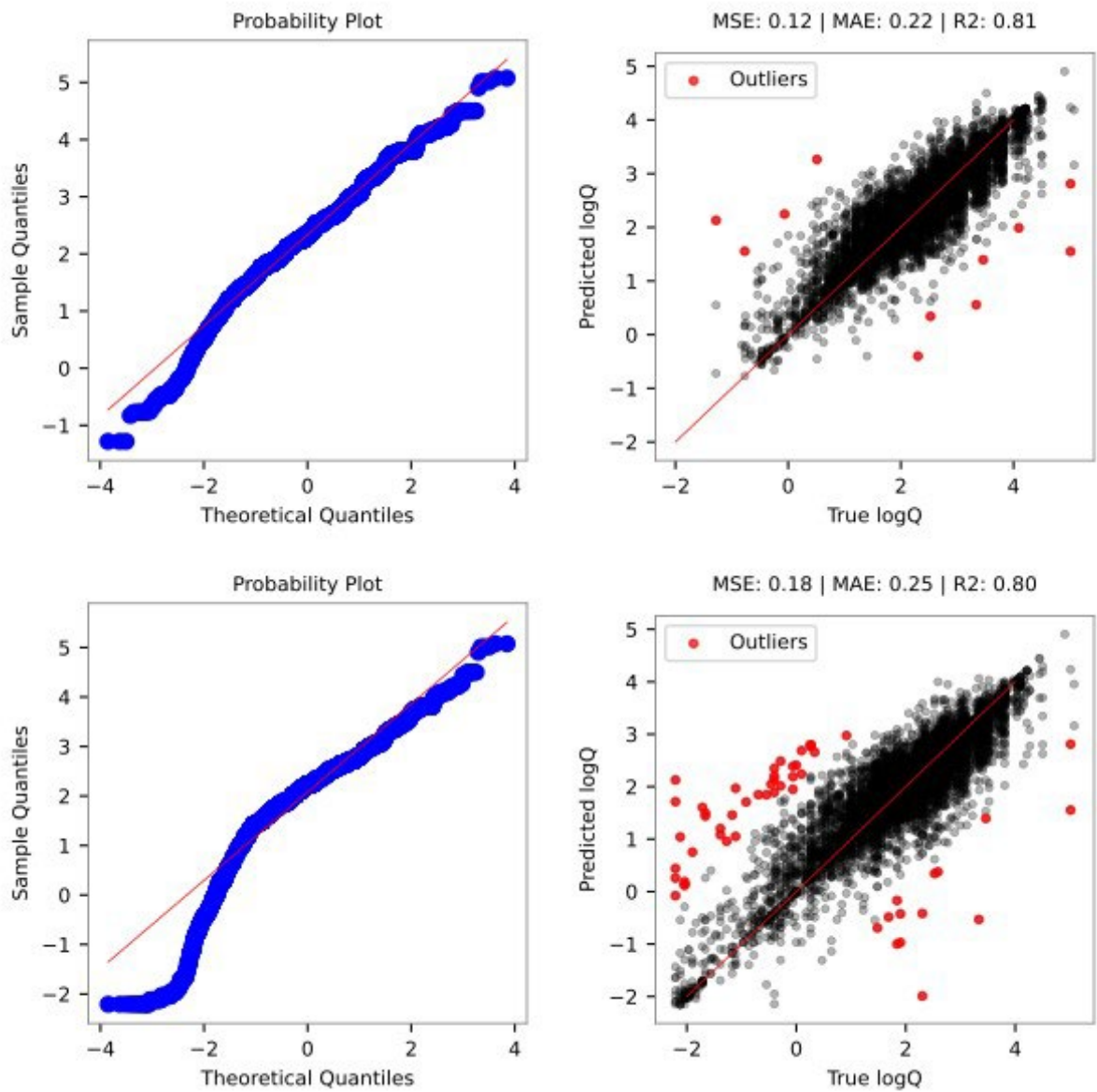

Appendix H. Comparing regression performance using various balancing techniques for the Extra Trees algorithm. From left to right, a) standard fit, b) binning continuous label and upsample low classes with SMOTE, c) linear prediction transformation

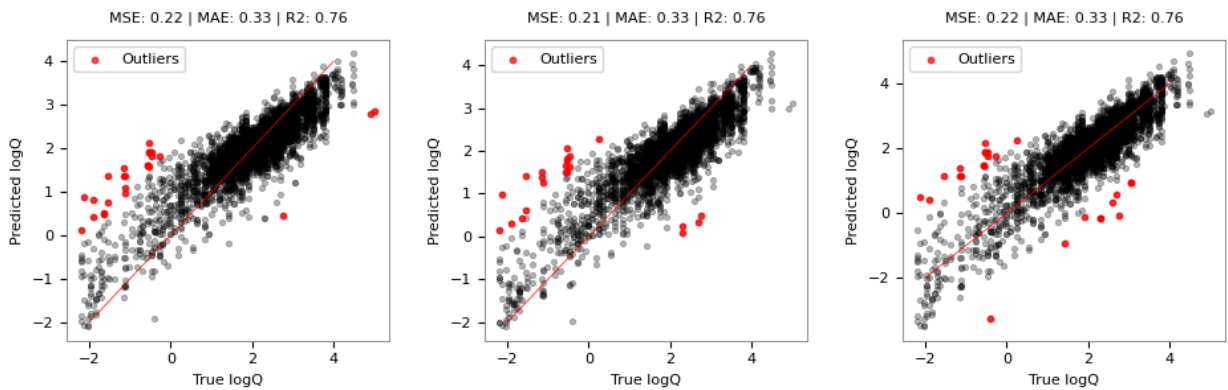